\documentclass[lettersize,journal]{IEEEtran}
\usepackage{amsmath,amsfonts}
\usepackage{algorithmic}
\usepackage{algorithm}
\usepackage{bbding}
\usepackage{array}
\usepackage[caption=false,font=normalsize,labelfont=sf,textfont=sf]{subfig}
\usepackage{textcomp}
\usepackage{stfloats}
\usepackage{url}
\usepackage{verbatim}
\usepackage{graphicx}
\usepackage{cite}
\usepackage{orcidlink}
\begin{document}

\title{EPIPTrack: Rethinking Prompt Modeling with Explicit and Implicit Prompts for Multi-Object Tracking}

\author{Yukuan Zhang\,\orcidlink{0000-0002-2811-0936}, Jiarui Zhao\,\orcidlink{0009-0008-5266-5043}, Shangqing Nie\,\orcidlink{0009-0003-8557-9313}, Jin Kuang\,\orcidlink{0009-0000-1907-8893}, Shengsheng Wang\,\orcidlink{0000-0002-8503-8061}
        \thanks{Yukuan Zhang, Jiarui Zhao, Shangqing Nie, and Shengsheng Wang are with the College of Computer Science and Technology, Jilin University, and also with the Key Laboratory of Symbolic Computation and Knowledge Engineering of Ministry of Education, Jilin University, Changchun 130012, China (e-mail: \{zyk24, zhaojr24, niesq25\}@mails.jlu.edu.cn; wss@jlu.edu.cn).}
        \thanks{Jin Kuang is Yangtze University, China (e-mail: gasking.stu@yangtzeu.edu.cn)}
        \thanks{Corresponding author: Shengsheng Wang.}
        }

\markboth{Journal of \LaTeX\ Class Files,~Vol.~14, No.~8, August~2021}%
{Shell \MakeLowercase{\textit{et al.}}: A Sample Article Using IEEEtran.cls for IEEE Journals}


\maketitle

\begin{abstract}
  Multimodal semantic cues, such as textual descriptions, have shown strong potential in enhancing target perception for tracking. However, existing methods rely on static textual descriptions from large language models, which lack adaptability to real-time target state changes and prone to hallucinations. To address these challenges, we propose a unified multimodal vision-language tracking framework, named EPIPTrack, which leverages explicit and implicit prompts for dynamic target modeling and semantic alignment. Specifically, explicit prompts transform spatial motion information into natural language descriptions to provide spatiotemporal guidance. Implicit prompts combine pseudo-words with learnable descriptors to construct individualized knowledge representations capturing appearance attributes. Both prompts undergo dynamic adjustment via the CLIP text encoder to respond to changes in target state. Furthermore, we design a Discriminative Feature Augmentor to enhance visual and cross-modal representations. Extensive experiments on MOT17, MOT20, and DanceTrack demonstrate that EPIPTrack outperforms existing trackers in diverse scenarios, exhibiting robust adaptability and superior performance.
\end{abstract}


\section{Introduction}
\IEEEPARstart{M}{ulti}-object tracking (MOT) is a fundamental task in computer vision, aiming to continuously localize multiple targets and maintain their identity consistency across video frames. It plays a critical role in a range of applications, including intelligent surveillance, autonomous driving\cite{hu2019joint}, and embodied intelligence\cite{khan2022m3t}. However, real-world challenges such as occlusion, target crowding, viewpoint variation, and uneven illumination greatly complicate identity consistency and demand greater robustness from tracking systems. To address these challenges, mainstream methods follow the tracking-by-detection (TBD) paradigm. In this framework, some approaches model target motion by introducing interpolation \cite{ocsort}, reconstruction \cite{strongsort}, and compensation strategies \cite{ucmctrack,sparsetrack} to mitigate issues such as trajectory fragmentation and drift. Notably, the conventional Kalman Filter \cite{kalman1960new} exhibits notable limitations when handling non-linear motion patterns. To this end, current studies propose alternatives such as neural Kalman Filters \cite{ikun}, noise-scale adaptive filters \cite{Giaotracker,Adaptrack}, and state-space-based architectures like Mamba \cite{xiao2024mambatrack,huang2024exploring}. 

Although motion modeling improves short-term association, it struggles to ensure reliable identity preservation in scenarios involving long-term occlusion or frequent interactions. To improve identity matching accuracy, some methods \cite{botsort,deepocsort,tracktrack,topic} integrate re-identification (ReID) modules that extract appearance features to enhance inter-object distinguishability.
\begin{figure}[t]
    \centering
    \includegraphics[width=1\columnwidth]{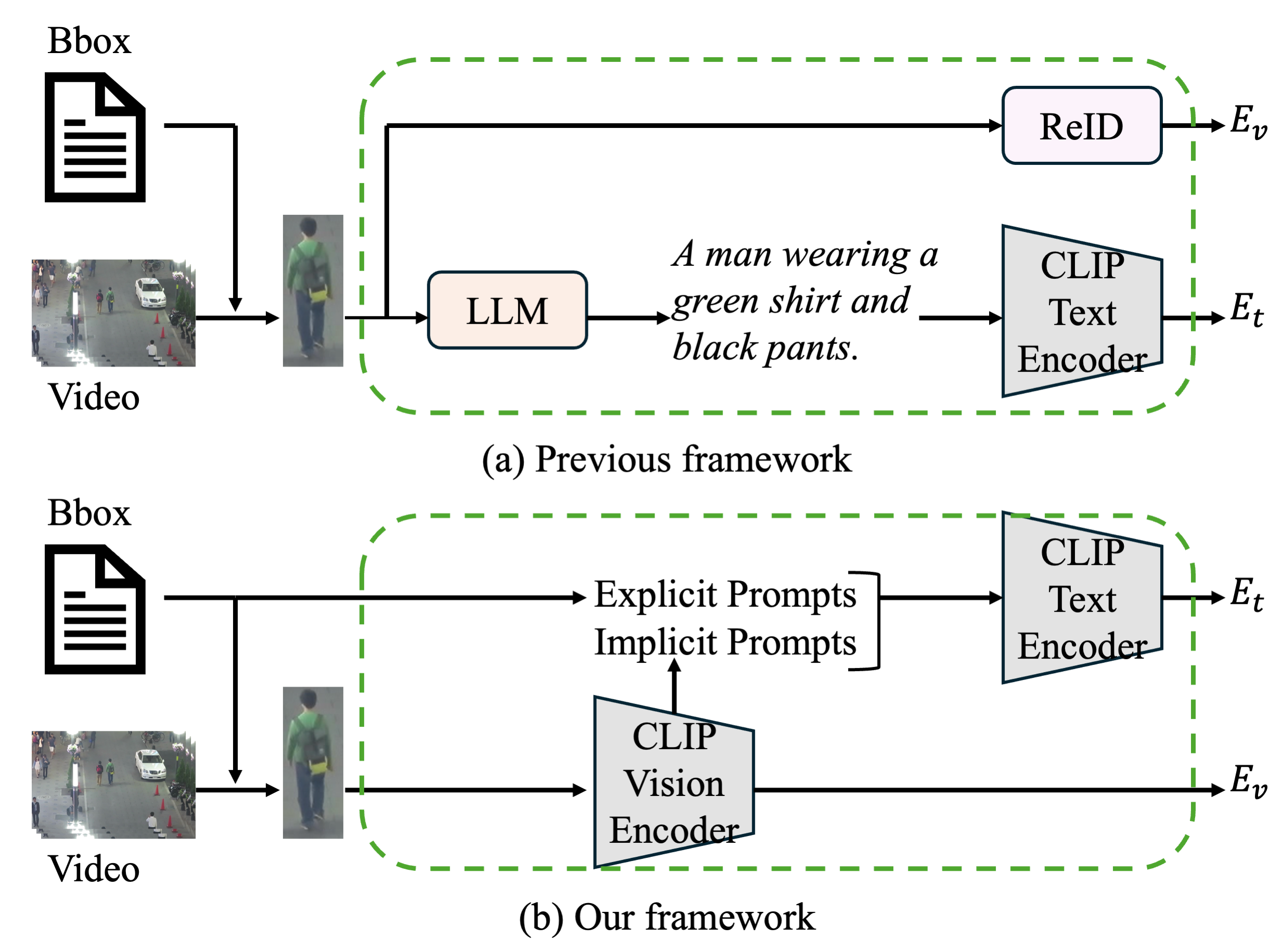} 
    \caption{Comparison between the proposed framework and mainstream framework.}
    \label{fig1}
    \end{figure}
    These methods show impressive progress. However, they frequently experience track loss and struggle to maintain long-term identity consistency. This highlights persistent limitations in semantic understanding and the modeling of long-range dependencies.

Recent studies \cite{li2025multi,ren2025semtg} integrate language modalities into MOT, enhancing the role of multimodal cues in addressing association ambiguities and semantic uncertainty. As shown in Fig. \ref{fig1}(a), such frameworks follow a modular architecture but suffer from several limitations. For instance, LGMOT \cite{li2025multi} relies on static textual descriptions to track dynamic targets, often causing semantic discrepancies and reducing tracking robustness. To improve adaptability, some methods \cite{ren2025semtg,li2025dynamic} employ dynamic language modules that update large language models (LLMs) \cite{hurst2024gpt} per frame to reflect target state changes. However, such mechanisms are prone to hallucinations inherent in LLMs, especially under occlusion. Furthermore, these frameworks typically comprise multiple heterogeneous modules (e.g., LLMs, text encoders, and ReID networks), resulting in high computational costs and impeding end-to-end optimization.

Extending these studies, we propose a novel vision-language tracking framework, as shown in Fig. \ref{fig1}(b).

Specifically, We utilize CLIP \cite{clip}, a pretrained vision-language model, as the backbone and incorporate explicit and implicit prompt mechanisms for task-specific tuning. For explicit prompts, we construct natural language descriptions based on salient motion cues, including detection score, speed, and depth. For example: “A person with identity 21 and a score of 0.85.” This description explicitly encodes both the identity and motion-related attributes of the target. CLIP employs hard prompt templates, such as “A photo of a [CLASS].” to achieve inter-class discrimination, supporting coarse-grained semantic modeling. In contrast, MOT tasks require fine-grained instance-level differentiation within the same category, making class-level representations suboptimal for precise instance discrimination. To overcome this limitation, we propose an implicit prompting strategy based on textual inversion \cite{gal2022image}, where a pseudo-word token is inserted into the text. Its semantic embedding is generated by the visual encoder, then refined within the text encoder. This allows the prompt to dynamically reflect temporal appearance variations while capture instance-level semantic attributes.

Unlike CLIP, our implicit prompt structure follows the format “[X]\textsubscript{1}[X]\textsubscript{2}[X]\textsubscript{3}...[X]\textsubscript{M} [PART] [$S^*$].”, where “$S^*$” is reserved for inserting the pseudo-word token that conveys individualized semantic cues for fine-grained target representation.

To further improve multimodal modeling, we propose a Discriminative Feature Augmentor that dynamically selects the Top-K most distinctive embeddings to enhance the fine-grained discriminative power of target representations. 

In summary, our main contributions are as follows:
\begin{itemize}
    \item To respond to target state changes, we rethink prompt modeling and propose explicit and implicit prompting methods. Without relying on LLMs, our approach enhances tracking stability and reliability.
    \item We propose a Discriminative Feature Augmentor to mine Top-K distinctive embeddings. It strengthens visual representations and guides the learning of textual representations in the latent space, thereby improving cross-modal modeling capability.
    \item We introduce a novel unified visual-language framework for MOT that operates without auxiliary modules such as LLMs or ReID, offering a streamlined and effective cross-modal tracking solution.
    \item Notably, the proposed method provides a plug-and-play design that seamlessly integrates into existing TBD paradigms. Extensive experiments on MOT17, MOT20, and DanceTrack validate its outstanding performance and achieve state-of-the-art results.
\end{itemize}
\begin{figure*}[t]
    \centering
    \includegraphics[width=2\columnwidth]{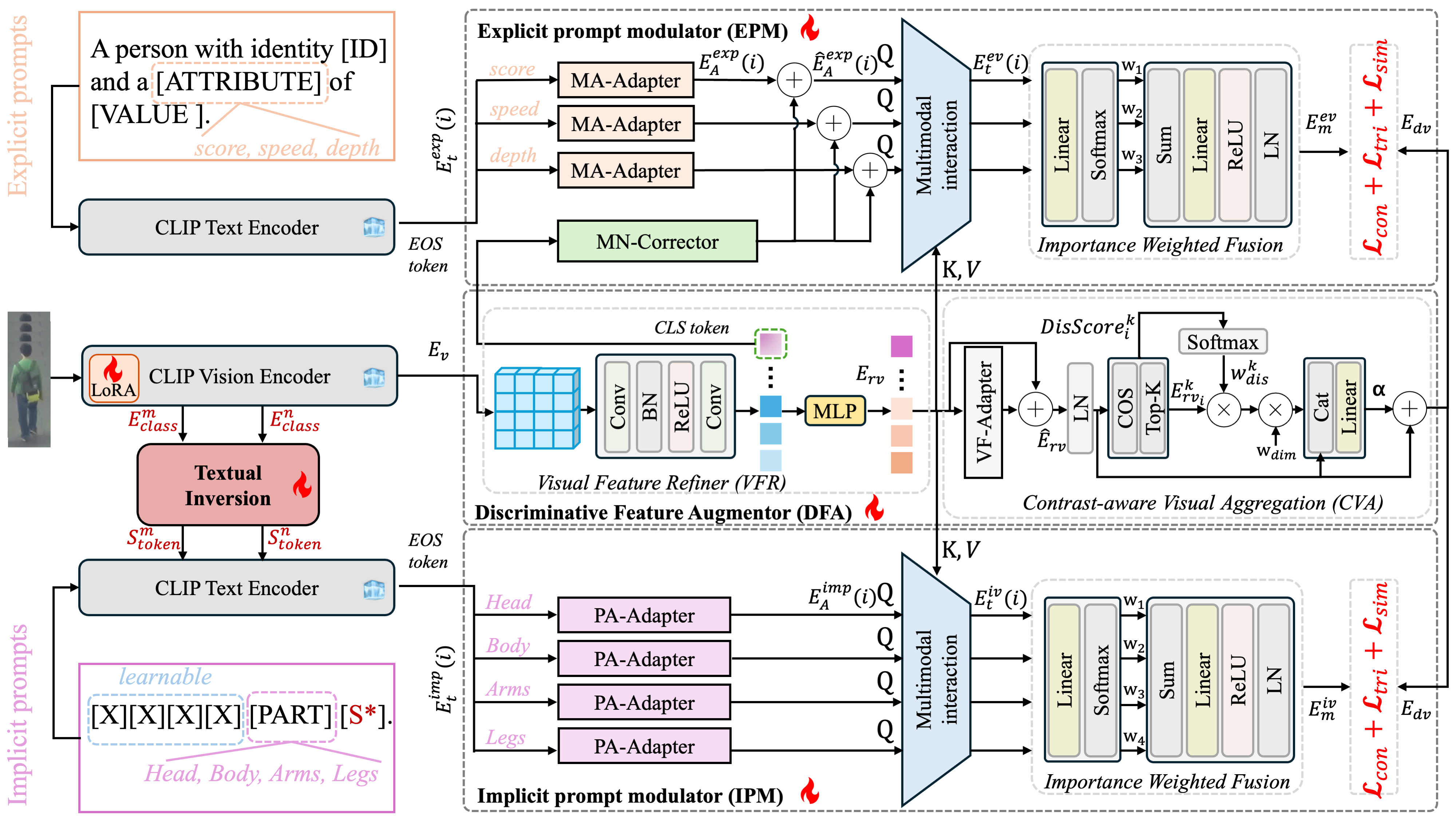} 
    \caption{A unified vision-language tracking framework, named EPIPTrack. MA and PA refer to motion attributes and body-part attributes, respectively. MN-Corrector is the motion noise corrector.}
    \label{fig2}
    \end{figure*}
\section{Related Work}
\subsection{Tracking-By-Detection}
In this paradigm, detectors localize targets in each frame, followed by cost matrix computation using Intersection over Union (IoU) or cosine similarity. Identity assignment is performed using the Hungarian Algorithm \cite{munkres1957algorithms}.

ByteTrack \cite{bytetrack} pioneers the use of low-confidence detections, breaking the reliance on high-confidence boxes. OC-SORT \cite{ocsort} emphasizes observation-driven association, UCMCTrack++ \cite{ucmctrack} introduces unified camera motion compensation, and SparseTrack \cite{sparsetrack} leverages pseudo-depth for enhanced spatial reasoning. These approaches prioritize motion modeling to improve localization stability. Appearance-based methods, such as BOT-SORT \cite{botsort}, TrackTrack \cite{tracktrack}, StrongSORT++ \cite{strongsort}, Deep OC-SORT \cite{deepocsort}, and Hybrid-SORT-ReID \cite{hybrid}, enhance long-term association by incorporating ReID modules. In contrast, our method adopts a vision-language multimodal strategy, using linguistic prompts to enrich target representation and boost association accuracy.
\subsection{Prompt learning}
In the downstream adaptation of vision–language models (e.g., CLIP\cite{clip}), conventional handcrafted templates are limited by their insufficient flexibility and generalization ability. Prompt learning has emerged as a parameter-efficient fine-tuning paradigm that replaces fixed templates with learnable prompts. CoOp\cite{coop} pioneered the use of continuous prompts in CLIP, achieving significant performance gains in few-shot scenarios; however, its static prompt design remains limited in generalization. Subsequently, numerous studies have focused on dynamic adaptation and cross-modal co-optimization. CoCoOp\cite{cocoop} leverages a meta-network to generate instance-specific dynamic prompts. MaPLe introduces cross-modal hierarchical prompt optimization. ProVP\cite{provp} enhances cross-layer prompt interactions within the visual encoder. CPL\cite{zhang2024concept} constructs a visual concept cache to generate dynamic prompts, further improving fine-grained visual classification. Moreover, chain-of-thought \cite{wei2022chain} prompting has substantially enhanced performance on complex tasks and logical reasoning by guiding multi-step inference. Despite notable progress in various downstream tasks, prompt learning remains underexplored in the dynamic visual domain of MOT. To address this gap, this paper builds upon CLIP as the backbone network and, considering the dynamic nature of MOT, proposes a dual-module architecture that integrates explicit and implicit prompts. Through their synergistic optimization, our approach enables efficient adaptation of CLIP to MOT tasks.
\subsection{Visual language tracking}
In recent years, language modality has advanced referring object tracking (ROT) \cite{wu2023referring,shao2024context,anh2024tp,nguyen2023type,feng2024memvlt,ma2024unifying} by facilitating cross-modal alignment, as seen in works like ZGMOT \cite{zgmot}. In contrast, this study emphasizes multimodal association. LGMOT \cite{li2025multi} uses LLMs to generate static attributes as fundamental semantic cues. SemTG-Track \cite{ren2025semtg} and DUTrack \cite{li2025dynamic} extend this approach to frame-wise generation for dynamic modeling, but introduce additional dependencies and risks of hallucination. LTrack \cite{yu2023generalizing} utilizes a handcrafted TrackBook, while IPMOT \cite{ipmot} proposes a learnable version. Both methods improve generalization in MOT but require a predefined number of targets during training, limiting adaptability. This work introduces explicit and implicit prompts to eliminate reliance on LLMs and enable dynamic modeling of target states, enhancing semantic association effectiveness.
\section{Method}

In this work, we propose a prompt learning strategy that adapts CLIP for downstream MOT. As shown in Fig. \ref{fig2}, we design a unified multimodal multi-object tracking framework, EPIPTrack. It builds on CLIP’s visual and language encoders, and introduces an Explicit Prompt Modulator, an Implicit Prompt Modulator, and a Discriminative Feature Augmentor to guide multimodal representation learning.
\subsection{Prompt description}
\textbf{Explicit Prompts}. For each target trajectory, we record the historical observation sequence $ O = [\text{ID}, f, x_1, y_1, x_2, y_2, s] $, where ID denotes the target identifier, $f$ is the timestamp, $ (x_1, y_1) $ and $ (x_2, y_2) $ represent the top-left and bottom-right coordinates of the bounding box, respectively, and $s$ indicates the confidence score. This sequence provides rich spatiotemporal motion information, forming the foundation for generating dynamic explicit prompts.

We observe that methods such as OC-SORT and Hybrid-SORT primarily leverage target speed, while SparseTrack and CAMOT \cite{CAMOT} focus more on depth information as a discriminative cue. Inspired by this, we incorporate both velocity and depth attributes to better capture spatiotemporal target dynamics. Additionally, since detection confidence score reflects visibility, typically higher when the target is fully exposed, we integrate it to enhance semantic perception.

Finally, the explicit prompt template takes the form: “A person with identity [ID] and a [ATTRIBUTE] of [VALUE].” It is used in real time for each target to characterize the temporal motion state. Speed and depth at time $f$ are defined in Eq. \ref{speed} and Eq. \ref{deep}, respectively.
\begin{equation}\label{speed}
    Speed_{f} = \sqrt{(w_{f} - w_{f-1})^2 + (h_{f} - h_{f-1})^2} 
\end{equation}
\begin{equation}\label{deep}
    Depth_{f} = H_{img} - y_{2_f} 
    \end{equation}
where $ h $ and $ w $ denote the height and width of the bounding box, respectively, and $ H $ is the height of the image.

\textbf{Implicit Prompts}. Originally developed for image synthesis, pseudo-tokens leverage textual inversion learning mechanism \cite{gal2022image} to encapsulate semantic concepts and fine-grained visual details from images. This cross-modal capacity aligns naturally with the demands of dynamic appearance modeling in MOT. Here, we propose the extension of pseudo-tokens to the MOT domain by introducing implicit prompts.

We design a lightweight Textual Inversion Network (TI-Net) that takes as input the global visual representation ($\mathbf{E}^{i}_{\text{CLS}} \in R^{768}$) extracted from the $i$-th layer of the visual encoder, and transforms it as follows:
\begin{equation}\label{tinework}
    \mathbf{S}^{i}_{\text{PSE}}=\text{Proj}(\text{MLP}(\mathbf{E}^{i}_{\text{CLS}}))\in R^{512}
\end{equation}

The resulting pseudo-token $\mathbf{S}^{i}_{\text{PSE}}$ is injected into the embedding space of the corresponding layer in the text encoder to capture appearance attributes.

In addition, we incorporate a soft prompt “[X]\textsubscript{1}[X]\textsubscript{2}[X]\textsubscript{3}...[X]\textsubscript{M}”, where $\text{X} \in R^{512}$ denotes a learnable text token. A single soft prompt is shared across all instances to encode coarse-grained category priors (e.g., person), capturing the general structure of human appearance. Unlike traditional handcrafted prompts (e.g., “a photo of”), the soft prompt is optimized end-to-end to learn transferable knowledge. This design helps suppress task-irrelevant features (e.g., background textures) and  improves discriminability during affinity measurement. Finally, our structured implicit prompt template is: “[X]\textsubscript{1}[X]\textsubscript{2}[X]\textsubscript{3}...[X]\textsubscript{M} [PART] [$S^*$].”, where $s^*$ indicates the pseudo-token position and PART denotes body parts such as head, body, arms, and legs. Without relying on additional body-part-level annotations, the placeholder [PART] encourages attention to different regions.
\subsection{Prompt modulator}
\begin{figure}[t]
    \centering
    \includegraphics[width=0.9\columnwidth]{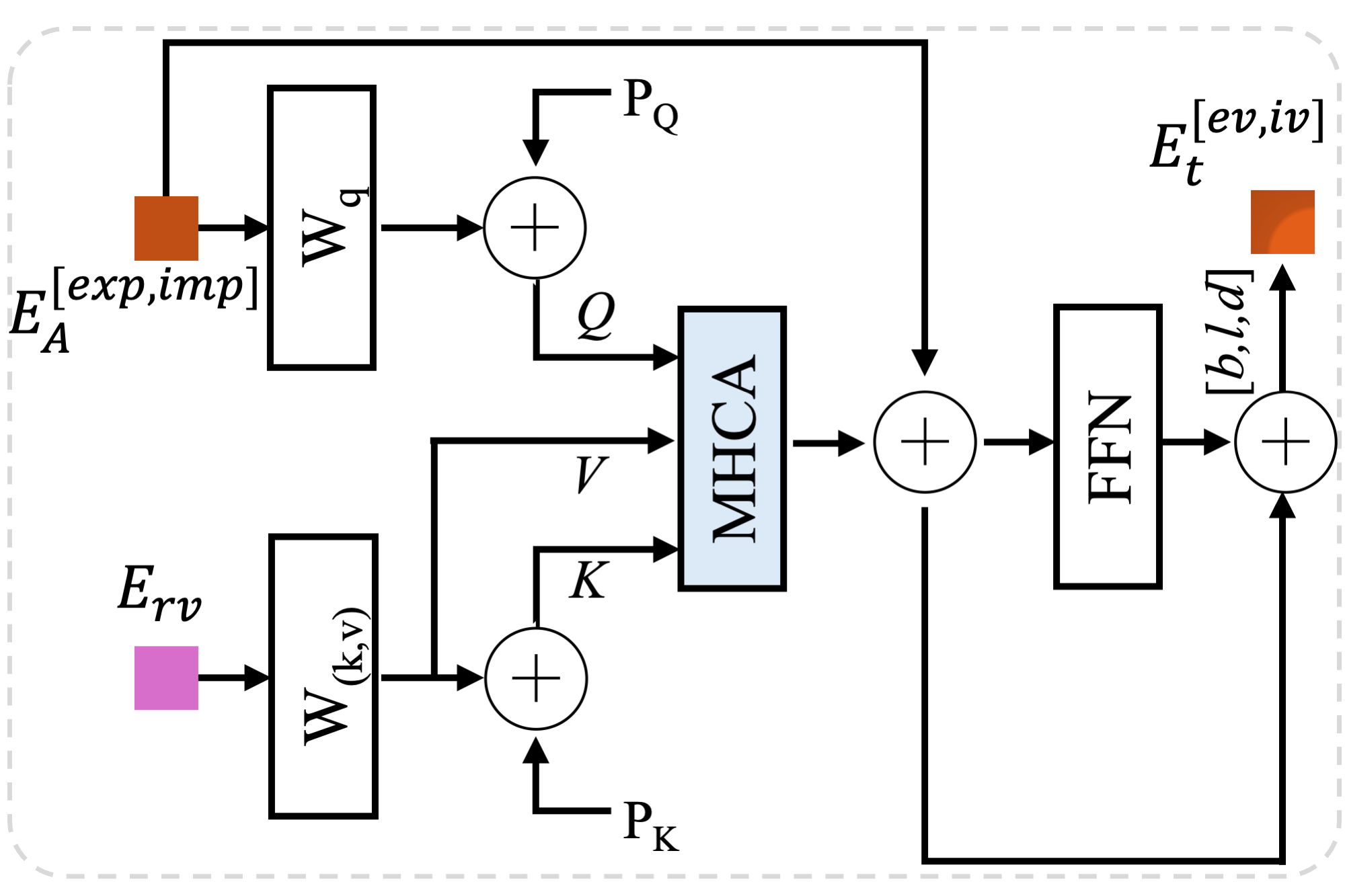} 
    \caption{Multimodal Interaction Network. We employ a single-layer Multi-Head Cross-Attention (MHCA). [$P_Q$,$P_K$] are learnable positional encodings.}
    \label{fig3}
\end{figure}
This module comprises an Attribute Adapter, a Multimodal Interaction Module, and an Importance-weighted Fusion Network, which collectively enhance the capacity for multimodal semantic modeling.

\textbf{Attribute Adapter}. The explicit prompt contains three sentences describing motion attributes: score, speed, and depth. The implicit prompt includes four sentences aligned with body parts: head, body, arms, and legs. These are encoded by CLIP into textual embeddings \{$\mathbf{E}^{exp}_{t}$,$\mathbf{E}^{imp}_{t}$\}. Attribute Adapters transform the EOS token via a lightweight linear layer, producing enriched representations \{$\mathbf{E}^{exp}_{A}$,$\mathbf{E}^{imp}_{A}$\} for multimodal modeling.

\textbf{Multimodal Interaction}. As shown in Fig. \ref{fig3}, $\mathbf{E}^{exp}_{A}\in R^{[b,3,d]}$ and $\mathbf{E}^{imp}_{A}\in R^{[b,4,d]}$ are input to the MHCA module with residual connections as queries ($Q$), while the refined visual feature $\mathbf{E}_{rv}$ acts as the key ($K$) and value ($V$). This enables the textual embeddings to integrate complementary visual cues, yielding cross-modally enhanced representations. Rather than using modality concatenation, we adopt a text-guided approach that enhances semantics to align visual and linguistic features, improving representational consistency. The final outputs are the text-guided multimodal embeddings \{$\mathbf{E}^{ev}_{t}\in R^{[b,3,d]}$,$\mathbf{E}^{iv}_{t}\in R^{[b,4,d]}$\}.

\textbf{Weighted Fusion}. In MOT scenarios, objects exhibit diverse appearance and motion patterns, influenced by dynamic factors such as interactions, occlusions, and scene crowding. These variations lead to unequal contributions of attribute subspaces, making equal weighting suboptimal for modeling individual differences. To address this, an importance-weighted fusion mechanism is designed to adaptively emphasizes the most informative attribute dimensions. The process can be formulated as follows:
\begin{equation}\label{weight fusion}
    \begin{gathered}
        \mathbf{E}^{x}_{t}=concat_{2}(\mathbf{E}^{x}_{t}(i))\in R^{[b,l,d]},\\
        att^{x} = \mathbf{W_{1}}\mathbf{E}^{x}_{t} \in R^{[b,l,1]},\\
        w_{i} = \frac{exp^{att^{x}_{i}}}{\sum_{i=1}^{l}exp^{att^{x}_{i}}},\\
        \mathbf{E}^{x}_{m} = \mathbf{W_{2}}\sum_{i=1}^{l}(w_{i}\times \mathbf{E}^{x}_{t}[:,i,:])\in R^{[b,1,d]},\\
        s.t. \ \ x \in \{ev,iv\}, l \in \{3, 4\}
    \end{gathered}
\end{equation}
where $\mathbf{W}_{1}\in R^{[d,1]}$ and $\mathbf{W}_{2}\in R^{[d,d]}$ denote the weight matrices of linear layers. This yields a unified multimodal representation composed of \{$\mathbf{E}^{ev}_{m}$, $\mathbf{E}^{iv}_{m}$\}. 

Observational noise \cite{zhang2025robust} may interfere with motion attribute modeling, leading to semantic shifts ($\epsilon_{\Delta}$). To mitigate noise-induced distortion, we introduce a Motion Noise Corrector ($\xi$) for explicit calibration. It comprises four Linear–ReLU–LayerNorm blocks, following a projection path of 512→1024→1024→512. The process is as follows:
\begin{equation}\label{corre noise}
    \begin{aligned}
        \hat{\mathbf{E}}^{exp}_{A}(i) &= \mathbf{E}^{exp}_{A}(i) - \epsilon_{\Delta}\\
        &=\mathbf{E}^{exp}_{A}(i) + \xi(\mathbf{E}^{cls}_{v})
    \end{aligned}
\end{equation}
\subsection{Feature augmentor}
In this section, we propose a Discriminative Feature Augmentor from the perspective of the visual modality. The initial visual embeddings $\mathbf{E}_{v}\in R^{[b,l,d]}$ are reshaped into a two-dimensional form $\mathbf{E}_{v}\in R^{[b\times l,d]}$, where $b$, $l$, and $d$ represent the batch size, sequence length, and dimension, respectively. To enhance inter-channel semantic and structural representation, a lightweight convolutional module is employed. This is followed by a bottleneck-style MLP that further refines the global features, yielding $\mathbf{E}_{rv}$. This representation serves as a critical input for subsequent multimodal interactions, with the refinement process ensuring semantic alignment for cross-modal consistency.

We introduce a contrastive-aware visual aggregation mechanism, where $\mathbf{E}_{rv}$ is processed by a Visual Feature Adapter (VF-Adapter). It consists of a single-layer linear mapping with a residual connection. The adapter captures inter-target structural differences to generate structurally-aware representations that support subsequent contrastive learning:
\begin{equation}\label{vfadapter}
    \hat{\mathbf{E}}_{rv} = \mathbf{E}_{rv} + \mathrm{LN}(\sigma(\mathbf{W}\mathbf{E}_{rv}))
\end{equation}

Each instance undergoes L2 normalization, followed by the computation of cosine distances between it and other targets. This process is defined as follows:
\begin{equation}\label{disscore}
    DisScore_{i,j}=\mathcal{D}_{cos}^{i,j}(\hat{\mathbf{E}}_{rv})=1 - \frac{\mathbf{\hat{E}}^i_{rv} \cdot \mathbf{\hat{E}}^j_{rv}}{\|\mathbf{\hat{E}}^i_{rv}\| \cdot \|\mathbf{\hat{E}}^j_{rv}\|}
\end{equation}

This results in a matrix $\mathbf{DisScore} \in R^{b \times b}$, where each entry $DisScore_{i,j}$ represents the cosine distance between the $i$-th and $j$-th targets. A higher score indicates lower semantic similarity. For each target, the $K$ most semantically dissimilar instances are selected as contrastive samples $\mathbf{E}^{K}_{rv_i}$, as shown in Eq. \ref{topk}. The model extracts differentiating information from these samples to enhance the target representation.
\begin{equation}\label{topk}
    \{DisScore^K_{i},\mathbf{E}^{K}_{rv_i}\} = \text{Top-K}_{j\ne i}(DisScore_{i,j},\mathbf{E}_{rv_j})
\end{equation}

To aggregate these contrastive features, the process operates at two levels. At the instance level, a distance-based softmax function assigns importance to the contrastive samples to emphasize more informative instances. At the channel level, a learnable scaling vector modulates the contribution of each dimension within the differentiating features, defined as:
\begin{equation}\label{weight1 fusion}
    \begin{gathered}
        w^{k}_{dis} = \frac{exp^{DisScore^{k}_{i}}}{\sum_{k=1}^{K}exp^{DisScore^{k}_{i}}},\\
        \mathbf{E}_{diff}=\mathbf{W}_{dim}\odot \sum^{K}_{k=1}(w^{k}_{dis}\times \mathbf{E}^{k}_{rv_i}),\\
        s.t.\ \mathbf{W}_{dim}\in R^d,initialized\ as\ 1
    \end{gathered}
\end{equation}

We concatenate the $\mathbf{E}_{diff}$ and $\mathbf{\hat{E}}{rv}$, followed by a linear layer with residual connection for feature co-optimization:
\begin{equation}\label{edv}
    \mathbf{E}_{dv} = \alpha \text{Linear}([\mathbf{E}_{diff},\mathbf{\hat{E}}_{rv}]) + \mathbf{\hat{E}}_{rv}
\end{equation}
where $\alpha$ is the coefficient hyperparameter.

By explicitly emphasizing semantic discrepancies across instances, the model learns more discriminative visual embeddings $\mathbf{E}_{dv}$ during representation learning.
\section{Training Objective}
To facilitate the learning of discriminative representations, we encourage the model to maximize inter-identity separation, reinforce intra-identity similarity, and ensure feature consistency across modalities. To this end, we introduce a supervised contrastive loss between the multimodal embeddings $\mathbf{E}^{x}_m$ and the visual embedding $\mathbf{E}_{dv}$ to enhance cross-modal consistency and identity discriminability. Specifically, given a batch of $N$ samples, let $\mathbf{m}_i$ and $\mathbf{v}_j$ denote the normalized embeddings drawn from \{$\mathbf{E}^{x}_m$, $\mathbf{E}_{dv}$\} corresponding to samples $i$ and $j$, respectively. The contrastive loss is formulated as follows:
\begin{equation}\label{speed}
    \mathcal{L}_{con} = -\frac{1}{N} \sum_{i=1}^N \log \frac{\sum_{j \in \text{Pos}(i)} \exp(\mathbf{m}_i \cdot \mathbf{v}_j / \tau)}{\sum_{k=1}^N \exp(\mathbf{m}_i \cdot \mathbf{v}_k / \tau)}
\end{equation}

In addition, a triplet loss is introduced to further enhance intra-class compactness and inter-class separability. It is computed bidirectionally from multimodal to visual and from visual to multimodal to better align the feature distributions across modalities. The formulation is as follows:
\begin{equation}
    \begin{gathered}
        \mathcal{L}^{tri}_{v2m,i}\!=\!\max\{\max_{j\!\in\!\text{Pos}(i)}\!d(\mathbf{v}_i,\!\mathbf{m}_j)\!-\!\min_{j\!\in\!\text{Neg}(i)}\!d(\mathbf{v}_i,\!\mathbf{m}_j)\!+\!\alpha,\!0\}\\
        \mathcal{L}^{tri}_{m2v,i}\!=\!\max\{\max_{j\!\in\!\text{Pos}(i)}\!d(\mathbf{m}_i,\!\mathbf{v}_j)\!-\!\min_{j\!\in\!\text{Neg}(i)}\!d(\mathbf{m}_i,\!\mathbf{v}_j)\!+\!\alpha,\!0\}\\
        \mathcal{L}_{tri} = \frac{1}{2N}\sum_{i=1}^{N}(\mathcal{L}^{tri}_{v2m,i}+\mathcal{L}^{tri}_{m2v,i})
    \end{gathered}
\end{equation}
where $d(\mathbf{X},\mathbf{Y}) = 1 - \frac{\mathbf{X}\cdot \mathbf{Y}}{\|\mathbf{X}\| \cdot \|\mathbf{Y}\|}$ denotes the cosine distance, and $\alpha$ is a margin hyperparameter set to $0.3$. 

We employ hard positive and negative mining to focus the training on the most challenging sample pairs.

A similarity distribution loss\cite{jiang2023cross} is introduced to refine feature structures. Specifically, it aligns the predicted similarity distributions from multimodal to visual and visual-to-multimodal with the ground-truth distribution defined by object identities. The ground-truth similarity distribution ($\mathbf{P}_{i,j}$) is defined as follows:
\begin{equation}
    \begin{gathered}
        \mathbf{L}_{i,j}=\begin{cases}
            1, & \text{if } \mathrm{id}_i = \mathrm{id}_j \\
            0, & \text{otherwise}
            \end{cases}\\
        \mathbf{P}_{i,j} = \frac{\mathbf{L}_{i,j}}{\sum_{k=1}^{N}(\mathbf{L}_{i,k})}
    \end{gathered}
\end{equation}

The predicted similarities are computed using cosine similarity between L2-normalized embeddings:
\begin{equation}
    \mathbf{Q}^{v2m}_{i,j}=\frac{\text{exp}((\mathbf{v}_{i}\cdot \mathbf{m}_{j})/ \tau)}{\sum_{k=1}^{N}\text{exp}((\mathbf{v}_{i}\cdot \mathbf{m}_{j})/ \tau)}
\end{equation}

The visual to multimodal loss for sample $i$ is:
\begin{equation}
    \mathcal{L}^{sim}_{v2m,i}=\sum_{j=1}^{N}\mathbf{Q}^{v2m}_{i,j}(\text{log}\mathbf{Q}^{v2m}_{i,j}-\text{log}(\mathbf{P}_{i,j}+\epsilon))
\end{equation}

Following the same procedure, we obtain $\mathcal{L}^{sim}_{m2v,i}$, where $\epsilon = 10^{-8}$ is used to prevent numerical instability. We define the similarity distribution loss as follows:
\begin{equation}
    \mathcal{L}_{sim} = \frac{1}{N}\sum_{i=1}^{N}(\mathcal{L}^{sim}_{v2m,i}+\mathcal{L}^{sim}_{m2v,i})
\end{equation}

The total loss is:
\begin{equation}
    \mathcal{L}_{all} = \mathcal{L}_{con} + \mathcal{L}_{tri}+ \mathcal{L}_{sim}
\end{equation}
\section{Experiments}
\subsection{Dataset and Evaluation Metric}
\textbf{Datasets}. We conduct experiments on three widely used datasets: MOT17 \cite{mot17}, MOT20 \cite{mot20}, and DanceTrack \cite{sun2022dancetrack}. MOT17 poses a range of real-world challenges such as camera motion, poor lighting, occlusion, and motion blur, making it ideal for testing robustness under diverse conditions. MOT20 focuses on extremely crowded scenes with an average of 170 pedestrians per frame, serving as a benchmark for high-density tracking. DanceTrack features dance performances with appearance similarity and complex, non-linear movements.

\textbf{Metrics}. We adopt HOTA \cite{hota} as the primary metric, providing a higher-order evaluation that jointly considers detection, association, and localization. Auxiliary metrics include IDF1 \cite{idf1} for identity preservation, MOTA \cite{mota} for detection accuracy, and AssA\cite{hota} for association accuracy.
\subsection{Implementation Details}
During training, target regions are cropped from the images based on ground-truth annotations and resized to (256, 128). Random cropping and horizontal flipping are applied for data augmentation. We adopt CLIP ViT-B/16 as the backbone. The text encoder is kept frozen, while the visual encoder is initialized with CLIP-ReID \cite{li2023clip} pretrained weights and subsequently fine-tuned on the MOT dataset using Low-Rank Adaptation (LoRA). The length $M$ of the learnable text token [X] in the implicit prompt template is set to $4$. In the Discriminative Feature Augmentor, the number of contrastive samples $K$ is set to $5$, as specified in Eq. \ref{topk}, and the hyperparameter $\alpha$ in Eq. \ref{edv} is set to $0.2$. We optimize the model using contrastive loss ($\mathcal{L}_{con}$), triplet loss ($\mathcal{L}_{tri}$), and similarity distribution loss ($\mathcal{L}_{sim}$) computed between multimodal embeddings ($\mathbf{E}^{x}_{m}$) and visual embeddings ($\mathbf{E}_{dv}$).

During the tracking process, trackers based on the TBD paradigm typically adopt a two-step association strategy. The first step computes the matching cost between confirmed trajectories and detection boxes using IoU or cosine similarity. The second step performs supplementary matching between unmatched trajectories (including those in lost or tentative states) and the remaining candidate detections. Our method is plug-and-play and can be seamlessly integrated into existing tracking frameworks. Specifically, we compute the cosine similarity between multimodal and visual embeddings to obtain a cost matrix, as defined in Eq. \ref{cos distance}. This matrix is utilized at two key stages. First, Track Reassociation (TR) is performed in the third step to match previously unmatched trajectories with detections. Second, Fusion Refinement (FR) is applied in the first step by averaging the cost matrix with the original similarity scores, thereby improving association accuracy.
\begin{equation}\label{cos distance}
  \mathcal{D}_{\cos}(\mathbf{E}^x_m, \mathbf{E}_{dv}) = \frac{1}{2} \sum_{x} \left( 1 - \frac{\mathbf{E}^x_m \cdot \mathbf{E}_{dv}}{\|\mathbf{E}^x_m\| \cdot \|\mathbf{E}_{dv}\|} \right)
\end{equation}
\subsection{Comparison with State-of-the-Art Methods}
We conduct a quantitative evaluation of the proposed method and compare it to state-of-the-art approaches. The results on the MOT17, MOT20, and DanceTrack test sets are presented in Table \ref{mot17}, \ref{mot20}, and \ref{DanceTrack}, respectively. Performance analysis indicates that EPIPTrack achieves leading performance across all datasets. Taking the MOT17 dataset with diverse challenges as an example, EPIPTrack achieves 67.2 HOTA and 83.2 IDF1, outperforming all mainstream methods based on either motion or appearance features.

In comparison with existing multimodal methods, LTrack and IPMOT adopt a query-based paradigm \cite{zeng2022motr}. However, these methods perform less effectively than EPIPTrack across multiple metrics. This performance gap may arise from an inherent conflict between learning multimodal semantics and achieving precise spatial localization. In contrast, both LGMOT and SemTG-Track adopt the same TBD paradigm and use YOLOX as the detector, enabling a more direct comparison with EPIPTrack. Experimental results consistently demonstrate the superiority of EPIPTrack in key metrics. For example, on the MOT20 dataset featuring dense crowds, it achieves 65.9 HOTA and 81.3 IDF1; on DanceTrack, which involves complex actions and non-linear motion, it attains 68.7 HOTA, 70.6 IDF1, and 93.4 MOTA.
\begin{table}[t]
    \centering
    \setlength{\tabcolsep}{3pt}
    \caption{Quantitative results on the MOT17 test set.}
    \begin{tabular}{ccccc}
    \hline
    Tracker                        &Ref.& HOTA$\uparrow$ & IDF1$\uparrow$ & MOTA$\uparrow$  \\ \hline
    \multicolumn{2}{l}{$Motion$-$Based$}\\
    ByteTrack\cite{bytetrack}                      &ECCV2022& 63.1 & 77.3 & 80.3  \\
    OC-SORT\cite{ocsort}                        &CVPR2023& 63.2 & 77.5 & 78.0  \\
    SparseTrack\cite{sparsetrack}                &TCSVT2025    & 65.1 & 80.1 & 81.0  \\
    UCMCTrack+\cite{ucmctrack}                   &AAAI2024  & 65.7 & 81.0 & 80.6  \\\hline
    \multicolumn{2}{l}{$Appearance$-$Based$} \\
    StrongSORT++\cite{strongsort}                 &TMM2023    & 64.4 & 79.5 & 79.6  \\
    Deep OC-SORT\cite{deepocsort}                  &ICIP2023 & 64.9 & 80.6 & 79.4  \\
    BOT-SORT\cite{botsort}                       &arXiv2022& 65.0   & 80.0 & 80.5     \\
    TOPICTrack\cite{cao2025topic}                       &TIP2025& 63.9   & 78.7 & 78.8     \\
    Hybrid-SORT-ReID\cite{hybrid}               &AAAI2024& 64.0   & 78.7 & 79.9     \\
    TrackTrack\cite{tracktrack}                  &CVPR2025   & 67.1 & 83.1 & 81.8 \\ \hline
    \multicolumn{2}{l}{$Vision$-$Language$-$Based$} \\
    LTrack\cite{yu2023generalizing}               &AAAI2023          & 57.5 & 69.1 & 72.1  \\
    IPMOT\cite{ipmot}                         &arXiv2024& 58.2 & 69.6 & 73.2  \\
    LGMOT\cite{li2025multi}                  &TCSVT2025       & 65.6 & 81.7 & 81.0      \\
    SemTG-Track\cite{ren2025semtg}              &ESWA2025      & 67.2 & 82.6 & \textbf{82.3}     \\
    \textbf{EPIPTrack(Ours)}                &-& \textbf{67.2} & \textbf{83.2} & 81.8  \\ \hline
    \end{tabular}
    \label{mot17}
\end{table}
\begin{table}[t]
    \centering
    \setlength{\tabcolsep}{3pt}
    \caption{Quantitative results on the MOT20 test set.}
    \begin{tabular}{ccccc}
        \hline
        Tracker                        &Ref.& HOTA$\uparrow$ & IDF1$\uparrow$ & MOTA$\uparrow$  \\ \hline
        \multicolumn{4}{l}{$Motion$-$Based$}                        \\
        ByteTrack\cite{bytetrack}        &ECCV2022& 61.3          & 75.2          & 77.8          \\
        OC-SORT\cite{ocsort}          &CVPR2023& 62.1          & 75.9          & 75.5          \\
        SparseTrack\cite{sparsetrack}   & TCSVT2025  & 63.4          & 77.3          & 78.2          \\
        UCMCTrack+\cite{ucmctrack}      &AAAI2024 & 62.8          & 77.4          & 75.6          \\\hline
        \multicolumn{4}{l}{$Appearance$-$Based$}                    \\
        StrongSORT++\cite{strongsort}    &TMM2023   & 62.6          & 77.0          & 73.8          \\
        Deep OC-SORT\cite{deepocsort}    &ICIP2023 & 63.9          & 79.2          & 75.6          \\
        BOT-SORT\cite{botsort}         &arXiv2022& 63.3          & 77.5          & 77.8          \\
        TOPICTrack\cite{cao2025topic}                       &TIP2025& 62.6   & 77.6 & 72.4     \\
        Hybrid-SORT-ReID\cite{hybrid}&AAAI2024 & 63.9          & 78.4          & 76.7          \\
        TrackTrack\cite{tracktrack}   &CVPR2025    & 65.7          & 80.9          & 78.0          \\ \hline
        \multicolumn{4}{l}{$Vision$-$Language$-$Based$}               \\
        LTrack \cite{yu2023generalizing}   &AAAI2023       & 46.8          & 61.1          & 57.8          \\
        IPMOT\cite{ipmot}          &arXiv2024 & 49.2          & 62.5          & 68.3          \\
        ZGMOT\cite{zgmot}          &arXIV2023 & 61.4          & 75.5          & 77.6          \\
        SemTG-Track\cite{ren2025semtg}   &ESWA2025  & 63.5          & 77.5          & \textbf{78.2} \\
        \textbf{EPIPTrack(Ours)} &-& \textbf{65.9} & \textbf{81.3} & 77.9          \\ \hline
        \end{tabular}
    \label{mot20}
\end{table}
\begin{table}[t]
    \centering
    \setlength{\tabcolsep}{3pt}
    \caption{Quantitative results on the DanceTrack test set.}
    \begin{tabular}{ccccc}
        \hline
        Tracker                        &Ref.& HOTA$\uparrow$ & IDF1$\uparrow$ & MOTA$\uparrow$  \\ \hline
        \multicolumn{4}{l}{$Motion$-$Based$}                        \\
        ByteTrack\cite{bytetrack}        &ECCV2022& 47.3          & 52.5          & 89.5          \\
        OC-SORT\cite{ocsort}          &CVPR2023& 55.1          & 54.9          & 92.2          \\
        SparseTrack\cite{sparsetrack}  &TCSVT2025    & 55.5          & 58.3          & 91.3          \\
        UCMCTrack+\cite{ucmctrack}      &AAAI2024 & 63.6          & 65.0          & 88.9          \\ \hline
        \multicolumn{4}{l}{$Appearance$-$Based$}                    \\
        StrongSORT++\cite{strongsort}    &TMM2023   & 55.6          & 55.2          & 91.1          \\
        TOPICTrack\cite{cao2025topic}                       &TIP2025& 58.3   & 58.4 & 90.9     \\
        Deep OC-SORT\cite{deepocsort}    &ICIP2023 & 61.3          & 61.5          & 92.3          \\
        Hybrid-SORT-ReID\cite{hybrid} &AAAI2024& 65.7          & 67.4          & 91.8          \\
        TrackTrack\cite{tracktrack}    &CVPR2025   & 66.5          & 67.8          & \textbf{93.6} \\ \hline
        \multicolumn{4}{l}{$Vision$-$Language$-$Based$}               \\
        IPMOT\cite{ipmot}           &arXiv2024& 61.9          & 62.0          & 88.2          \\
        LGMOT\cite{li2025multi}      &TCSVT2025     & 61.8          & 60.5          & 89.0          \\
        \textbf{EPIPTrack(Ours)}  &-& \textbf{68.7} & \textbf{70.6} & 93.4          \\ \hline
        \end{tabular}
    \label{DanceTrack}
\end{table}
\begin{table}[t]
    \centering
    \caption{Equal-weighted overall scores per sequence.}
    \begin{tabular}{cccc}
        \hline
        Tracker                        & HOTA$\uparrow$ & IDF1$\uparrow$ & AssA$\uparrow$  \\ \hline
        \multicolumn{4}{l}{$MOT17$}                                                          \\
        TrackTrack & 60.0                     & 74.8                     & 61.7                     \\
        EPIPTrack  & 60.4                     & 75.6                     & 62.5                     \\ \hline
        \multicolumn{4}{l}{$MOT20$}                                                          \\
        TrackTrack & 62.3                     & 76.6                     & 63.3                     \\
        EPIPTrack  & 62.9                     & 77.2                     & 64.3            \\ \hline
        \multicolumn{4}{l}{$DanceTrack$}                                                     \\
        TrackTrack & 66.6                     & 69.3                     & 53.8                     \\
        EPIPTrack  & 68.8            & 72.1            & 57.2                     \\ \hline
        \end{tabular}
    \label{Average comparison}
\end{table}

Experimental results indicate that, without relying on LLMs, EPIPTrack effectively models multimodal consistency via a CLIP-driven unified vision-language framework. This framework comprehensively leverages target-specific attributes and demonstrates significant advantages in maintaining long-term discriminability.

We observe that variations in video sequence length and target count in the test set may unevenly influence final scores, affecting fair performance comparison. To mitigate this, we conducted additional experiments assigning equal weights to each sequence, as shown in Table \ref{Average comparison}. Focusing on tracker performance, EPIPTrack significantly outperforms the baseline TrackTrack, further confirming the effectiveness of the language modality in enhancing semantic understanding.
\subsection{Ablation Study}
\begin{table}[!t]
    \centering
    \caption{Ablation study on association strategy.}
    \begin{tabular}{cccc}
        \hline
        Method      & HOTA$\uparrow$ & IDF1$\uparrow$ & AssA$\uparrow$ \\ \hline
        \multicolumn{4}{l}{$MOT17$}                        \\
        TrackTrack & 69.1 & 85.1 & 72.8 \\
+TR        & 69.2 & 85.4 & 73.1 \\
+FR        & 69.7 & 86.2 & 74.1 \\ \hline
        \multicolumn{4}{l}{$DanceTrack$}                    \\
        TrackTrack & 63.4 & 66.9 & 49.9 \\
+TR        & 64.3 & 68.6 & 51.2 \\
+FR        & 65.9 & 70.9 & 53.9 \\  \hline
        \end{tabular}
    \label{MOT Ablation experiment}
\end{table}
\begin{table}[!t]
    \centering
    \caption{Zero-Shot Generalization on DanceTrack-Val (Trained on MOT17). To isolate the impact of the vision-language model, the motion-based Hybrid-SORT is used as the baseline.}
    \begin{tabular}{cccc}
        \hline
        Method     & HOTA$\uparrow$ & IDF1$\uparrow$ & AssA$\uparrow$ \\ \hline
        Hybrid-SORT & 50.4 & 54.6 & 35.7 \\
        +TR        & 50.5 & 55.0 & 35.9 \\
        +FR        & 50.9 & 55.3 & 36.6 \\ \hline
        \end{tabular}
    \label{Zero shot}
\end{table}
\begin{table}[!h]
    \centering
    \caption{Comparison of Runtime and Accuracy for ByteTrack Variants on MOT17-Val (A100 GPU).}
    \begin{tabular}{cccc}
        \hline
        Method    & HOTA$\uparrow$  & IDF1$\uparrow$  & FPS$\uparrow$   \\ \hline
        Byte      & 74.17 & 83.51 & 412.1 \\
        Byte+ReID & 74.69 & 83.61 & 10.0  \\
        Byte+EPIP & 74.94 & 84.73 & 6.7   \\ \hline
        \end{tabular}
    \label{Time expense}
\end{table}
\begin{table}[!t]
    \centering
    \caption{Universal compatibility of EPIP with motion and appearance trackers.}
    \begin{tabular}{cccc}
        \hline
        Tracker      & HOTA$\uparrow$ & IDF1$\uparrow$ & AssA$\uparrow$ \\ \hline
        \multicolumn{4}{l}{$Motion$-$Based$}                        \\
        ByteTrack    & 74.2 & 83.5 & 74.8 \\
        +EPIP       & 74.9 & 84.7 & 76.4 \\
        Hybrid-SORT  & 73.8 & 82.4 & 74.0 \\
        +EPIP       & 74.3 & 83.2 & 74.9 \\
        OC-SORT      & 72.8 & 81.1 & 72.4 \\
        +EPIP       & 73.5 & 81.9 & 73.5 \\ \hline
        \multicolumn{4}{l}{$Appearance$-$Based$}                    \\
        StrongSORT++   & 68.6 & 81.1 & 73.4 \\
        +EPIP       & 68.8 & 81.6 & 73.8 \\
        BOT-SORT     & 74.7 & 83.6 & 75.0 \\
        +EPIP       & 75.7 & 84.6 & 76.9 \\
        Deep OC-SORT & 72.9 & 81.5 & 73.6 \\
        +EPIP       & 73.2 & 82.2 & 74.1 \\ \hline
        \end{tabular}
    \label{plug and play }
\end{table}
\textbf{Effectiveness of association strategy}. Table \ref{MOT Ablation experiment} presents the performance contributions of the proposed association enhancements (TR and FR) on the MOT17 and DanceTrack validation sets. The results demonstrate that TR effectively mitigates missed associations and improves the recovery of fragmented trajectories. Building on this, FR enhances association accuracy by incorporating vision-language similarity during the initial processing stage. The combination yields consistent performance gains across both benchmarks, with substantially greater improvements on DanceTrack, where targets exhibit rapid motion and high visual similarity.

\textbf{Zero Shot}. Although DanceTrack provides relatively favorable scenarios for detection, its complex motion patterns and frequent target interactions pose significant challenges for association modeling. We evaluate the generalization capability of our vision-language framework on this previously unseen and challenging dataset. As shown in Table \ref{Zero shot}, integrating the TR and FR modules leads to notable improvements (+0.7 IDF1, +0.9 AssA) over the baseline, demonstrating that the learned cross-modal representations retain strong discriminative power under a distribution shift.

\textbf{Plug-and-play capability}. As shown in Table \ref{plug and play }, EPIP acts as a plug-and-play module that can be integrated into various trackers based on motion and appearance, consistently improving performance. These results highlight its versatility and demonstrate that incorporating language modality helps overcome the limitations of visual perception.
\subsection{Analysis of Inference Time and Accuracy}
Table \ref{Time expense} compares the computational efficiency of EPIP and traditional ReID modules. Compared to motion-based trackers, appearance-based and multimodal approaches generally require more computation. However, under comparable time overhead, EPIP delivers more substantial performance gains than ReID. For example, it achieves a $1.23$ improvement in IDF1. This indicates that EPIP offers a more efficient trade-off between accuracy and speed, while enabling richer modeling of semantic representations.
\subsection{Qualitative Analysis}
\begin{figure}[!h]
    \centering
    \includegraphics[width=1\columnwidth]{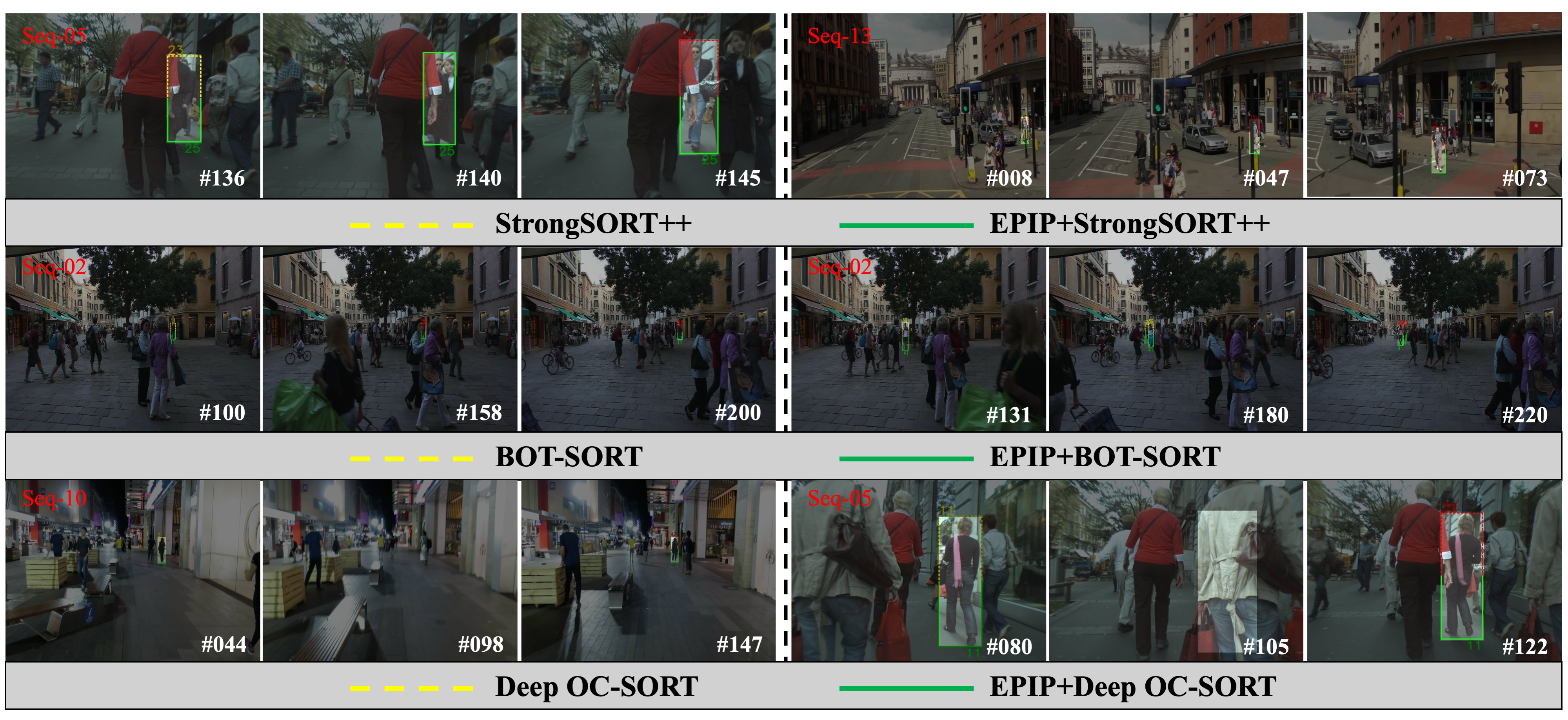} 
    \caption{Robustness evaluation in complex scenarios. Identity interruptions are indicated by color changes. The evaluated target is highlighted for clarity.}
    \label{fig4}
\end{figure}

Fig. \ref{fig4} demonstrates the robustness of EPIP when integrated with mainstream appearance-based trackers under complex conditions. In scenes with camera motion and occlusion (e.g., StrongSORT++ on Seq-05 and Seq-10), EPIP effectively mitigates ID switches and track fragmentation. For low-resolution targets and inter-object interactions (e.g., BOT-SORT on Seq-02), EPIP alleviates misassociation and tracking drift. Moreover, under distant and low-light settings (e.g., Deep OC-SORT on Seq-10), EPIP maintains consistent identity assignment, highlighting its effective vision-language fusion and enhanced identity discrimination.
\section{Extended Experiment}
This section aims to validate the effectiveness of each module and the overall rationality of the framework. The MOT17 training set is split into two subsets, one for training and the other for evaluation. We train the model using SGD with cosine learning rate decay and a constant warm-up phase to stabilize convergence. Key parameters include a batch size of $32$, an initial learning rate of $2 \times 10^{-5}$, a warm-up learning rate of $2 \times 10^{-6}$, and $100$ training epochs. All experiments are conducted on a single NVIDIA A100 GPU with 40GB VRAM. 

We adopt a threshold-wise multimodal similarity evaluation protocol to evaluate the discriminability of target representations under varying cosine similarity thresholds, reporting metrics including Precision and F1-score, as detailed below:
\begin{equation}
    \begin{gathered}
        \text{Precision}_{thr} = \frac{\text{TP}_{thr}}{\text{TP}_{thr}+\text{FP}_{thr}} \\
        \text{F1}_{thr} = 2 \cdot \frac{\text{Precision}_{thr} \cdot \text{Recall}_{thr}}{\text{Precision}_{thr} + \text{Recall}_{thr}}
    \end{gathered}
\end{equation}
where $\text{Recall}_{thr} = \frac{\text{TP}_{thr}}{\text{TP}_{thr}+\text{FN}_{thr}}$. 
\begin{table*}[!t]
    \centering
    \caption{Analysis of the effectiveness of loss terms.}
    \begin{tabular}{ccc|cc|cc|cc}
    \hline
    \multicolumn{3}{c|}{Loss} & \multicolumn{2}{c|}{$Thr$@$0.6$} & \multicolumn{2}{c|}{$Thr$@$0.7$} & \multicolumn{2}{c}{$Thr$@$0.8$} \\
    $\mathcal{L}_{con}$   & $\mathcal{L}_{tri}$   & $\mathcal{L}_{sim}$   & Pre$\uparrow$          & F1$\uparrow$           & Pre$\uparrow$          & F1$\uparrow$           & Pre$\uparrow$          & F1$\uparrow$           \\ \hline
    \CheckmarkBold&        &        & 0.44         & 0.61         & 0.63         & 0.77         & 0.84         & 0.91         \\
    \CheckmarkBold      & \CheckmarkBold      &        & 0.55         & 0.71         & 0.75         & 0.86         & 0.93         & 0.97         \\
    \CheckmarkBold      &        & \CheckmarkBold      & 0.45         & 0.62         & 0.71         & 0.83         & 0.93         & 0.96         \\
    \CheckmarkBold      & \CheckmarkBold      & \CheckmarkBold      & 0.55         & 0.71         & 0.79         & 0.88         & 0.96         & 0.98         \\ \hline
    \end{tabular}
    \label{tab:loss_ablation}
\end{table*}
\begin{table*}[!t]
    \centering
    \caption{This section presents an analysis of the effectiveness of the proposed modules, including the Explicit Prompt Modulator (EPM), Implicit Prompt Modulator (IPM), Contrastive-aware Visual Aggregation (CVA), Visual Feature Refiner (VFR), and Text Inversion Network (TI-Net).}
    \begin{tabular}{ccccc|cc|cc|cc|cc|cc}
        \hline
        \multicolumn{5}{c|}{Models} & \multicolumn{2}{c|}{$Thr$@$0.5$} & \multicolumn{2}{c|}{$Thr$@$0.6$} & \multicolumn{2}{c|}{$Thr$@$0.7$} & \multicolumn{2}{c|}{$Thr$@$0.8$} & \multicolumn{2}{c}{Consistency} \\
        EPM & IPM & CVA & VFR & TI-Net & Pre$\uparrow$        & F1$\uparrow$         & Pre$\uparrow$        & F1$\uparrow$         & Pre$\uparrow$        & F1$\uparrow$         & Pre$\uparrow$        & F1 $\uparrow$        & Gap$\downarrow$       & Align$\downarrow$     \\ \hline
        \CheckmarkBold   & \CheckmarkBold  &     &     &        & 0.00       & 0.00       & 0.00       & 0.00       & 0.00       & 0.00       & 0.00       & 0.00       & 1.08      & 1.74     \\
        \CheckmarkBold   & \CheckmarkBold & \CheckmarkBold   &     &        & 0.58       & 0.73       & 0.88       & 0.93       & 0.99       & 0.35       & 0.00       & 0.00       & 0.19      & 0.82     \\
        \CheckmarkBold   & \CheckmarkBold &     & \CheckmarkBold   &        & 0.86       & 0.92       & 0.98       & 0.81       & 0.50       & 0.24       & 0.00       & 0.00       & 0.46      & 0.71     \\
        \CheckmarkBold   & \CheckmarkBold & \CheckmarkBold   & \CheckmarkBold   &        & 0.35       & 0.52       & 0.58       & 0.73       & 0.83       & 0.91       & 0.97       & 0.99       & 0.05      & 0.16     \\
        \CheckmarkBold   & \CheckmarkBold  & \CheckmarkBold   & \CheckmarkBold   & \CheckmarkBold      & 0.37       & 0.54       & 0.60       & 0.75       & 0.84       & 0.91       & 0.98       & 0.99       & 0.04      & 0.16 \\ \hline    
        \end{tabular}
    \label{tab:model_ablation}
\end{table*}

Let $thr \in \{0.5, 0.6, 0.7, 0.8\}$ denote the threshold. For a given threshold $thr$, a matched pair of $\mathbf{E}^x_m$ and $\mathbf{E}_{dv}$ is classified as a true positive (TP) if their similarity exceeds $thr$; otherwise, it is classified as a false negative (FN). An unmatched pair with similarity exceeding $thr$ is classified as a false positive (FP). Unless otherwise specified, all reported results are based on the average metrics between $\mathbf{E}^{ev}_m$ and $\mathbf{E}_{dv}$, as well as between $\mathbf{E}^{iv}_m$ and $\mathbf{E}_{dv}$.
\subsection{Loss ablation}
Table \ref{tab:loss_ablation} systematically evaluates the impact of each loss component on multimodal representation learning. Using only the contrastive loss $\mathcal{L}_{con}$, the model achieves a precision of $0.44$ and an F1 score of $0.61$ at $thr$@$0.6$, improving to $0.84$ and $0.91$ at $thr$@$0.8$, indicating basic cross-modal alignment. However, without intra-class compactness constraints, it struggles to capture complex multimodal relations and distinguish challenging samples.

The introduction of triplet loss $\mathcal{L}_{tri}$ significantly improves performance, yielding a $10\%$ increase in F1 score at $thr$@$0.6$ and a $9\%$ gain in precision at $thr$@$0.8$. This highlights its effectiveness in promoting intra-class compactness and inter-class separability. The similarity distribution loss $\mathcal{L}_{sim}$, which aligns predicted similarities with identity priors via KL divergence, further refines global structure. While effective at high thresholds, it is slightly less robust than $\mathcal{L}_{tri}$ in handling ambiguous cases under low thresholds.

Combining all three losses yields the best overall performance, outperforming any single or partial configuration. This demonstrates the effectiveness of joint optimization and the complementary nature of the components.
\subsection{Module ablation}
Table \ref{tab:model_ablation} systematically evaluates the effectiveness of the proposed module in enhancing modal discriminability and cross-modal consistency. Precision and F1 score are adopted to assess discriminative capability across varying similarity thresholds. Additionally, Modality Gap and Alignment Score are introduced as complementary metrics to quantitatively measure consistency between each matched pair, as defined below:
\begin{equation}
    \begin{gathered}
        \text{Modality Gap} = \| \frac{1}{N_{test}}\sum_{i=1}^{N_{test}}\mathbf{m}_i-\frac{1}{N_{test}}\sum_{i=1}^{N_{test}}\mathbf{v}_i \|^{2}_{2}\\
        \text{Alignment}=\frac{1}{N_{test}}\sum_{i=1}^{N_{test}}\|\mathbf{m}_i-\mathbf{v}_i\|^{2}_{2}
    \end{gathered}
\end{equation}

The evaluation is performed across all test samples. The former captures the discrepancy at the cluster level, whereas the latter quantifies the alignment quality of each positive pair in the embedding space.

CLIP is adopted as the vision-language backbone, with lightweight prompt tuning applied via the Explicit and Implicit Prompt Modulators (EPM and IPM). However, training the prompt modulators alone is insufficient to adapt the model to the downstream MOT task (see the first row of Table \ref{tab:model_ablation}), likely due to domain gaps between the pretraining data and the tracking scenario. Integrating the Contrastive-aware Visual Aggregation (CVA) and Visual Feature Refiner (VFR) modules enables the model to extract task-relevant knowledge more effectively, thereby activating the latent capacity of the prompt modulators. Both modules yield notable performance gains under low-threshold settings (e.g., $thr$@$0.6$), indicating their effectiveness in improving sample discriminability.

However, under high-threshold settings (e.g., $thr$@$0.8$), using CVA or VFR alone yields limited performance gains, revealing suboptimal intra-class compactness that hampers discriminative capability. It is worth noting that VFR outperforms CVA in discriminative modeling, while the latter excels in cross-modal consistency modeling, making the two modules complementary. When CVA and VFR work jointly with EPM and IPM (as shown in the fourth row), the overall model performance is significantly enhanced. This collaboration improves intra-class compactness and enhances cross-modal consistency without compromising inter-class separability. 

Building on the above, the incorporation of a pseudo-token mechanism (last row) further improves model performance. Although primarily designed for implicit prompt modeling, its contribution to overall effectiveness is non-negligible. In summary, the results validate the effectiveness of our unified vision-language tracking framework and provide insights into improving cross-modal robustness and discriminability.
\subsection{Injection positions of pseudo-token}
Table \ref{tab:layer_ablation} presents an analysis of pseudo-token injection positions, where tokens are transferred from visual encoder layers into the text encoder via TI-Net. The results show that injecting pseudo-tokens into all layers of the text encoder does not lead to performance improvement, likely due to unstable optimization caused by the over-parameterization of injected information. In contrast, injecting only into deeper layers enhances the representational capacity of pseudo-tokens but yields smaller gains than mid-layer injection. Notably, the best performance is achieved when pseudo-tokens are injected into the 5th and 8th layers. This configuration is therefore adopted in our final design.
\begin{table}[!t]
    \centering
    \caption{Determine the injection position of pseudo-word tokens.}
    \begin{tabular}{ccc}
        \hline
                    & \multicolumn{2}{c}{$Thr$@$0.8$} \\
        Layer       & Pre$\uparrow$        & F1$\uparrow$         \\ \hline
        \{2-11\}      & 0.972      & 0.986      \\
        \{9, 10, 11\} & 0.973      & 0.986      \\
        \{2, 6, 10\}  & 0.975      & 0.987      \\
        \{2, 6\}      & 0.976      & 0.988      \\
        \{5, 8\}      & 0.977      & 0.988      \\ \hline
        \end{tabular}    
    \label{tab:layer_ablation}
\end{table}
\begin{table}[!t]
    \centering
    \caption{Comparison between wighted fusion and other fusion strategies.}
    \begin{tabular}{c|cc|cc|cc}
        \hline
                        & \multicolumn{2}{c|}{$Thr$@$0.6$} & \multicolumn{2}{c|}{$Thr$@$0.7$} & \multicolumn{2}{c}{$Thr$@$0.8$} \\
        Method          & Pre$\uparrow$        & F1$\uparrow$         & Pre$\uparrow$        & F1$\uparrow$         & Pre$\uparrow$        & F1$\uparrow$         \\ \hline
        SA & 0.39       & 0.56       & 0.62       & 0.77       & 0.88       & 0.94       \\
        Cat.   & 0.51       & 0.68       & 0.74       & 0.85       & 0.93       & 0.96       \\
        Wei.        & 0.55       & 0.71       & 0.79       & 0.88       & 0.96       & 0.98       \\ \hline
        \end{tabular}
    \label{tab:wighted_ablation}
\end{table}
\subsection{Alternative strategies for weighted fusion}
\begin{figure}[!t]
    \centering
    \includegraphics[width=1\columnwidth]{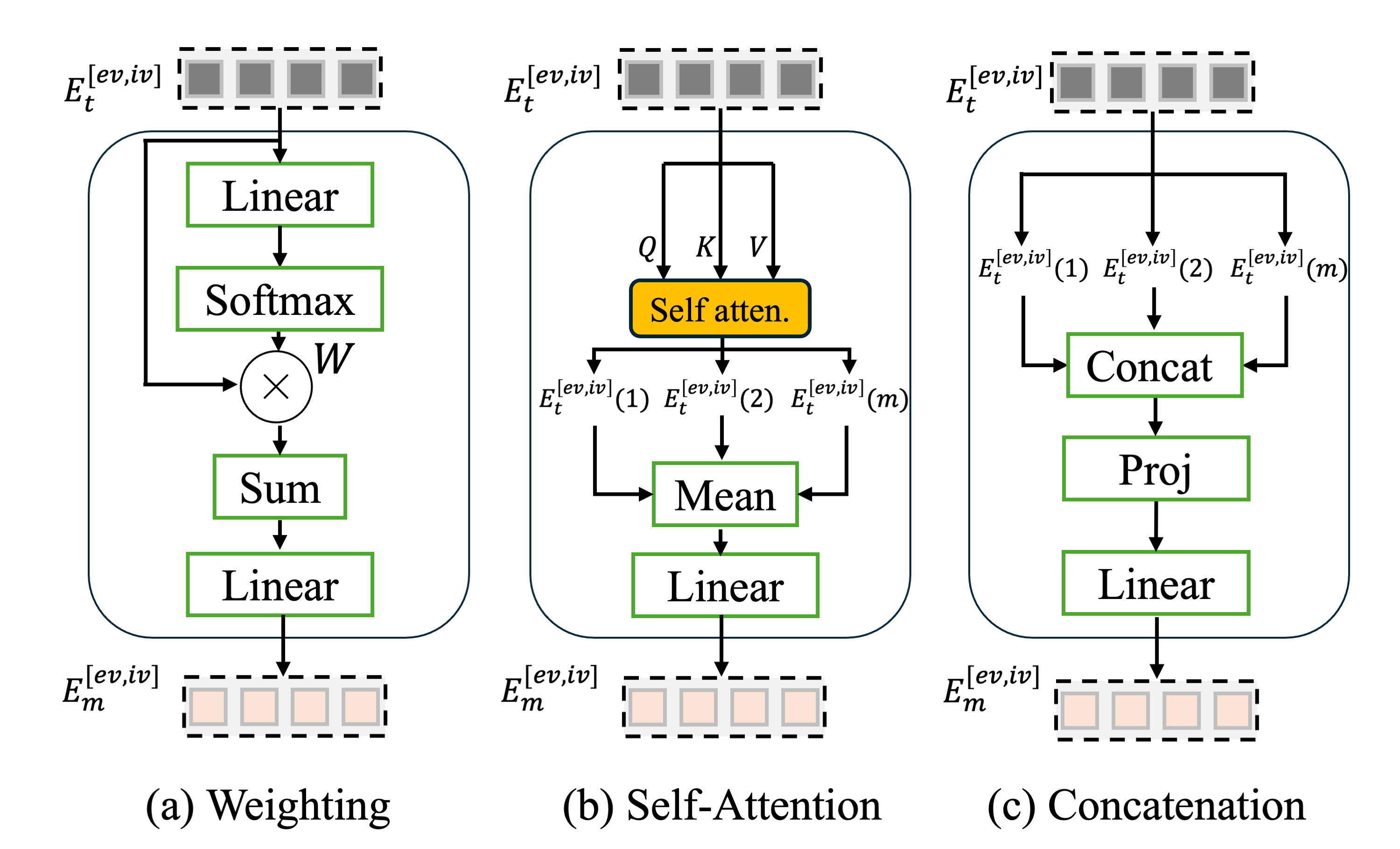} 
    \caption{Comparison of attribute fusion variants.}
    \label{fig5}
\end{figure}
\begin{figure}[!t]
    \centering
    \includegraphics[width=1\columnwidth]{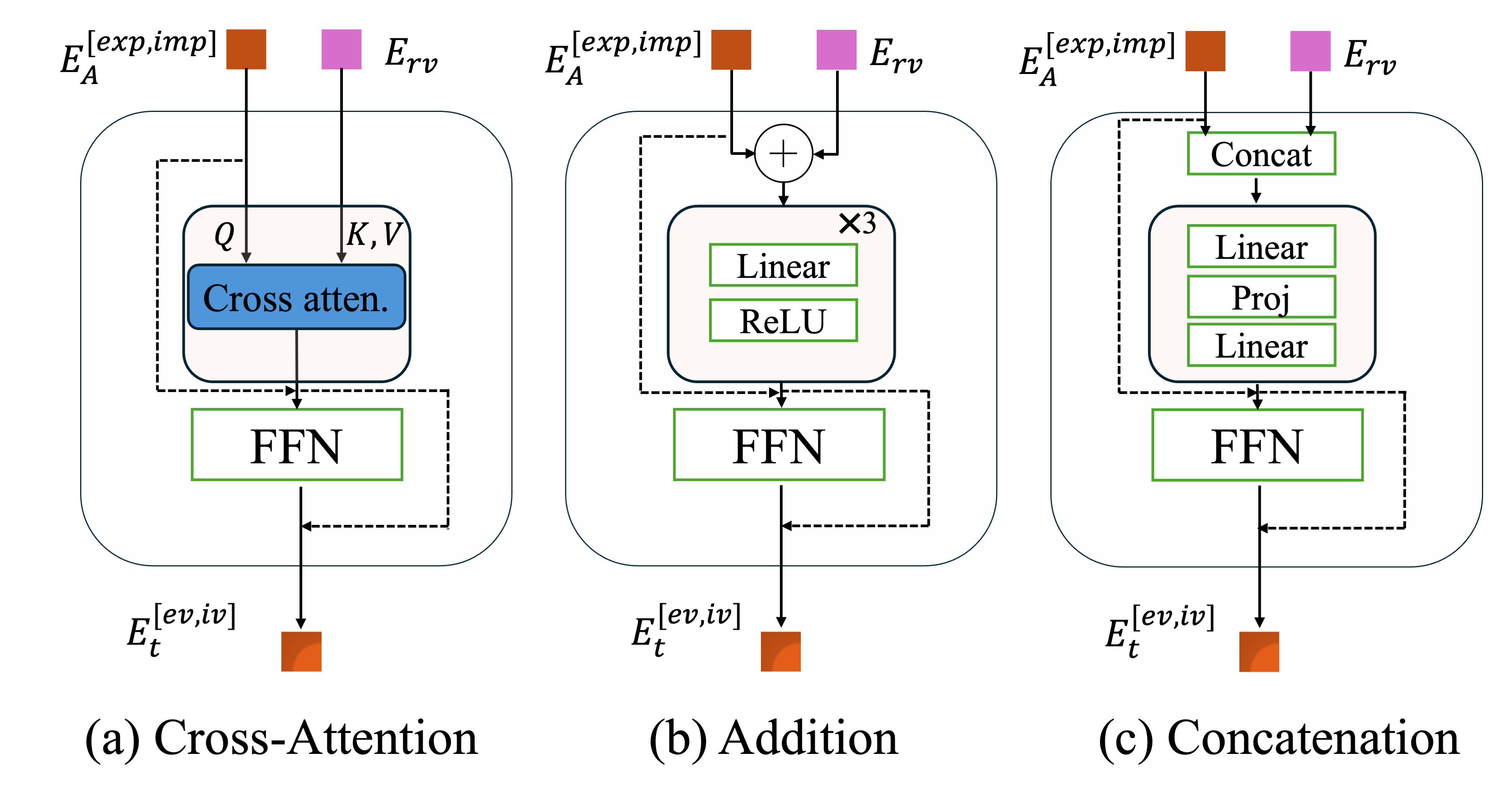} 
    \caption{Comparison of multimodal interaction variants.}
    \label{fig6}
\end{figure}
We explore various attribute fusion strategies and ultimately adopt the weighted fusion scheme illustrated in Fig. \ref{fig5}(a). Specifically, the self-attention pipeline processes attribute features to generate enhanced representations, followed by mean aggregation. In contrast, the concatenation pipeline stacks all attributes along the channel dimension and applies a linear projection to restore the original feature space.

Table \ref{tab:wighted_ablation} evaluates the performance of these three strategies. The results show that the self-attention approach yields unsatisfactory results, likely due to numerous trainable parameters introduced at the final stage, which increases optimization difficulty. Additionally, the mean aggregation operation may diminish critical discriminative features, thereby weakening the representational capacity of the model. The simple concatenation strategy achieves suboptimal performance by enabling coarse-grained integration of attribute information, though its contribution to final recognition accuracy is limited. In comparison, the weighted fusion method achieves the best performance by explicitly modeling the importance of different attributes. It learns to emphasize key attributes and suppress redundancy, enhancing the discriminability of the fused representation.
\begin{table}[!t]
    \centering
    \caption{Design of a multimodal interaction strategy.}
    \begin{tabular}{c|cc|cc|cc}
        \hline
                        & \multicolumn{2}{c|}{$Thr$@$0.6$} & \multicolumn{2}{c|}{$Thr$@$0.7$} & \multicolumn{2}{c}{$Thr$@$0.8$} \\
        Method          & Pre$\uparrow$        & F1$\uparrow$         & Pre$\uparrow$        & F1$\uparrow$         & Pre$\uparrow$        & F1$\uparrow$         \\ \hline
        Cat.   & 0.55       & 0.71       & 0.79       & 0.88       & 0.96       & 0.98       \\
        Add.        & 0.49       & 0.66       & 0.76       & 0.86       & 0.95       & 0.98       \\
        CA & 0.56       & 0.72       & 0.80       & 0.89       & 0.96       & 0.98       \\ \hline
        \end{tabular}
           
    \label{tab:multimodal_ablation}
\end{table}
\begin{figure}[!t]
    \centering
    \includegraphics[width=1\columnwidth]{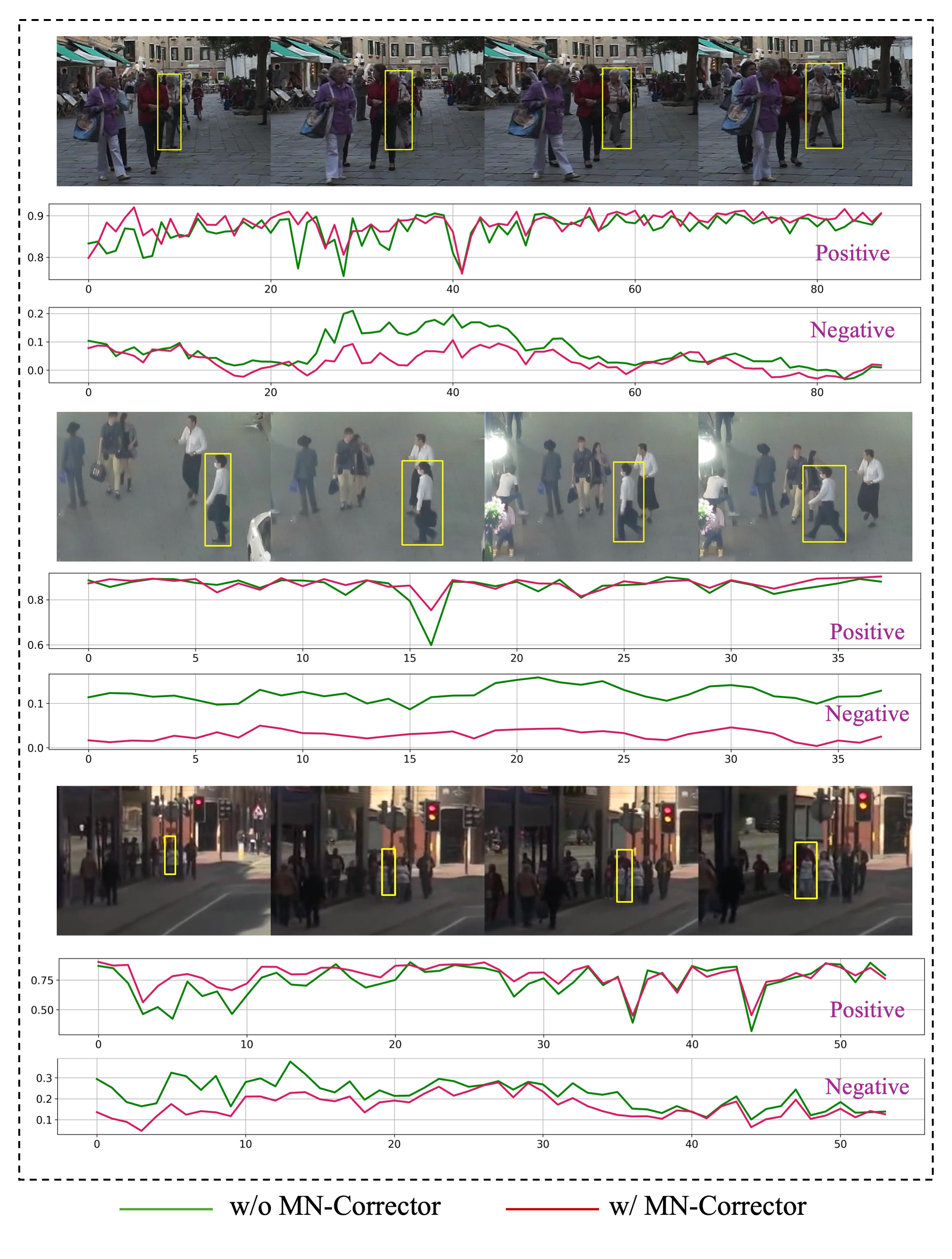} 
    \caption{Analysis of the practical performance of the Motion Noise Corrector.}
    \label{fig7}
\end{figure}
\subsection{Alternative multimodal interaction designs}
As a bridge between visual and textual modalities, the multimodal interaction module plays a pivotal role in enhancing the discriminability and semantic alignment of cross-modal representations. As illustrated in Fig. \ref{fig6}, we design three fusion strategies. Despite their structural differences, all share a unified design principle: using the textual modality as the residual stream, into which visual features are injected to facilitate semantic enrichment and cross-modal coordination.

Among them, the single-layer cross-attention mechanism treats textual features as queries, integrating visual context into each textual token. This design effectively models semantic relevance and yields highly discriminative multimodal representations (Table \ref{tab:multimodal_ablation}, last row). In contrast, the direct addition strategy employs a three-layer linear network to learn fused features. While it achieves competitive results at high thresholds, its performance deteriorates significantly at lower thresholds. The concatenation strategy fuses features via a linear layer, restores dimensionality through projection, and refines the representation with an additional linear transformation. Its performance is comparable to that of cross-attention (first row). However, it remains marginally inferior in overall effectiveness.

Given this comparative analysis, we select the cross-attention mechanism as the core design of our multimodal interaction module.
\begin{table}[!t]
    \centering
    \caption{Ablation study on the effects of explicit and implicit prompts. The combination of both yields the best performance across all metrics.}
    \begin{tabular}{cccc}
        \hline
        Method & HOTA$\uparrow$ & IDF1$\uparrow$ & AssA$\uparrow$ \\ \hline
        - & 74.9 & 84.7 & 76.4 \\
        w/o Imp. & 74.8 & 84.6 & 76.2 \\
        w/o Exp. & 74.6 & 84.0 & 75.7 \\ \hline
    \end{tabular}
    \label{emp_imp}
\end{table}
\subsection{Effect of combining explicit and implicit prompts}
The results in Table \ref{emp_imp} demonstrate that combining explicit and implicit prompts yields the best performance in MOT, as the former captures dynamic behavioral patterns and the latter offers stable appearance cues. This integration enhances target discriminability and identity preservation by generating more informative dynamic prompts.
\begin{table}[!t]
    \centering
    \caption{Computational breakdown on $1920 \times 1080$ resolution. An average of 67 pedestrians per frame.}
    \begin{tabular}{ccc}
        \hline
        Component                         & Time(ms)     & \% of Total    \\ \hline
        EPM         & 55.5         & 32.7          \\
        IPM         & 39.7         & 23.4          \\
        CVA & 1.0          & 0.6           \\
        MN-Corrector            & 0.5          & 0.3           \\
        VFR            & 0.6          & 0.4           \\
        Other                  & 72.6         & 42.7          \\ \hline
        \textbf{Total}                    & \textbf{170} & \textbf{100\%} \\ \hline
        \end{tabular}
    \label{computationla breakdown}
\end{table}
\subsection{MN-Corrector effectiveness}
Motion attributes in MOT are prone to observation noise, leading to semantic drift and reduced cross-modal discriminability. To mitigate this, we introduce the Motion Noise Corrector (MN-Corrector), which operates on the channel dimension of motion attribute features. By adaptively modulating channel-wise responses, it suppresses noise and aligns motion semantics. Such quantitative results are illustrated in Fig. \ref{fig7}, MN-Corrector improves feature similarity among positive samples and reduces confusion among negatives, thereby enhancing inter-class separability. This fine-grained adjustment enables accurate capture of critical motion information, and enhances the quality of cross-modal representations.
\subsection{Limitation and future work}
EPIPTrack exhibits superior cross-modal representation learning capabilities and achieves strong performance in MOT. It can be seamlessly integrated with existing TBD methods, significantly enhancing their modeling of target attribute information. However, the approach also presents certain limitations: it struggles to achieve high-precision association when relying solely on visual-language cues. This issue is similar to that faced by traditional appearance-based tracking methods, where spatial information remains indispensable for maintaining association accuracy. A potential solution to this limitation is to develop spatiotemporal trajectory modeling strategies to enhance the independent tracking capability of EPIPTrack. 

While the language modality offers a novel semantic cue that improves tracking performance, it also introduces additional computational overhead. We break down the runtime cost of each module, as detailed in Table \ref{computationla breakdown}. The proposed prompt modulation mechanism accounts for a substantial portion of the overall cost, due to dynamic temporal adjustment of instance-level textual descriptions. Moreover, the CLIP encoder built upon the ViT architecture is also a time-consuming component (included under “Other”). Exploring sparse computation strategies based on token-level similarity may help avoid redundant overhead. This work focuses on establishing a foundational unified vision-language tracking framework, and we will optimize the inference efficiency in future work.

\section{Conclusion}
In this work, we propose a unified multimodal vision-language tracking framework, EPIPTrack, built upon the CLIP foundation model. By incorporating explicit and implicit prompting mechanisms, the framework dynamically adapts to variations in target motion and appearance, enabling real-time state awareness. Unlike existing methods that rely on static textual descriptions or large language models, this approach operates without additional language model support, thereby mitigating issues such as model hallucination. Additionally, a discriminative feature enhancement module is designed to improve the consistency and discriminability of visual and language modalities. Extensive experiments demonstrate the superiority of EPIPTrack in joint vision-language tracking. The framework also offers strong plug-and-play capability, allowing seamless integration into existing TBD paradigms. It provides a more robust and scalable solution for MOT.
\section*{Acknowledgments}
This work is supported by the National Natural Science Foundation of China (62376106), The Science and Technology Development Plan of Jilin Province (20250102212JC).
\bibliographystyle{IEEEtran}
\bibliography{EPIPTrack}

\begin{thebibliography}{10}
\providecommand{\url}[1]{#1}
\csname url@samestyle\endcsname
\providecommand{\newblock}{\relax}
\providecommand{\bibinfo}[2]{#2}
\providecommand{\BIBentrySTDinterwordspacing}{\spaceskip=0pt\relax}
\providecommand{\BIBentryALTinterwordstretchfactor}{4}
\providecommand{\BIBentryALTinterwordspacing}{\spaceskip=\fontdimen2\font plus
\BIBentryALTinterwordstretchfactor\fontdimen3\font minus \fontdimen4\font\relax}
\providecommand{\BIBforeignlanguage}[2]{{%
\expandafter\ifx\csname l@#1\endcsname\relax
\typeout{** WARNING: IEEEtran.bst: No hyphenation pattern has been}%
\typeout{** loaded for the language `#1'. Using the pattern for}%
\typeout{** the default language instead.}%
\else
\language=\csname l@#1\endcsname
\fi
#2}}
\providecommand{\BIBdecl}{\relax}
\BIBdecl

\bibitem{hu2019joint}
H.-N. Hu, Q.-Z. Cai, D.~Wang, J.~Lin, M.~Sun, P.~Krahenbuhl, T.~Darrell, and F.~Yu, ``Joint monocular 3d vehicle detection and tracking,'' in \emph{Proceedings of the IEEE/CVF International Conference on Computer Vision}, 2019, pp. 5390--5399.

\bibitem{khan2022m3t}
M.~Khan, J.~Abu-Khalaf, D.~Suter, and B.~Rosenhahn, ``M3t: Multi-class multi-instance multi-view object tracking for embodied ai tasks,'' in \emph{International Conference on Image and Vision Computing New Zealand}.\hskip 1em plus 0.5em minus 0.4em\relax Springer, 2022, pp. 246--261.

\bibitem{ocsort}
J.~Cao, J.~Pang, X.~Weng, R.~Khirodkar, and K.~Kitani, ``Observation-centric sort: Rethinking sort for robust multi-object tracking,'' in \emph{Proceedings of the IEEE/CVF conference on computer vision and pattern recognition}, 2023, pp. 9686--9696.

\bibitem{strongsort}
Y.~Du, Z.~Zhao, Y.~Song, Y.~Zhao, F.~Su, T.~Gong, and H.~Meng, ``Strongsort: Make deepsort great again,'' \emph{IEEE Transactions on Multimedia}, vol.~25, pp. 8725--8737, 2023.

\bibitem{ucmctrack}
K.~Yi, K.~Luo, X.~Luo, J.~Huang, H.~Wu, R.~Hu, and W.~Hao, ``Ucmctrack: Multi-object tracking with uniform camera motion compensation,'' in \emph{Proceedings of the AAAI conference on artificial intelligence}, vol.~38, no.~7, 2024, pp. 6702--6710.

\bibitem{sparsetrack}
Z.~Liu, X.~Wang, C.~Wang, W.~Liu, and X.~Bai, ``Sparsetrack: Multi-object tracking by performing scene decomposition based on pseudo-depth,'' \emph{IEEE Transactions on Circuits and Systems for Video Technology}, 2025.

\bibitem{kalman1960new}
R.~E. Kalman, ``A new approach to linear filtering and prediction problems,'' 1960.

\bibitem{ikun}
Y.~Du, C.~Lei, Z.~Zhao, and F.~Su, ``ikun: Speak to trackers without retraining,'' in \emph{Proceedings of the IEEE/CVF Conference on Computer Vision and Pattern Recognition}, 2024, pp. 19\,135--19\,144.

\bibitem{Giaotracker}
Y.~Du, J.~Wan, Y.~Zhao, B.~Zhang, Z.~Tong, and J.~Dong, ``Giaotracker: A comprehensive framework for mcmot with global information and optimizing strategies in visdrone 2021,'' in \emph{Proceedings of the IEEE/CVF International conference on computer vision}, 2021, pp. 2809--2819.

\bibitem{Adaptrack}
K.~Shim, K.~Ko, J.~Hwang, and C.~Kim, ``Adaptrack: Adaptive thresholding-based matching for multi-object tracking,'' in \emph{2024 IEEE International Conference on Image Processing (ICIP)}.\hskip 1em plus 0.5em minus 0.4em\relax IEEE, 2024, pp. 2222--2228.

\bibitem{xiao2024mambatrack}
C.~Xiao, Q.~Cao, Z.~Luo, and L.~Lan, ``Mambatrack: a simple baseline for multiple object tracking with state space model,'' in \emph{Proceedings of the 32nd ACM International Conference on Multimedia}, 2024, pp. 4082--4091.

\bibitem{huang2024exploring}
H.-W. Huang, C.-Y. Yang, W.~Chai, Z.~Jiang, and J.-N. Hwang, ``Exploring learning-based motion models in multi-object tracking,'' \emph{arXiv e-prints}, pp. arXiv--2403, 2024.

\bibitem{botsort}
N.~Aharon, R.~Orfaig, and B.-Z. Bobrovsky, ``Bot-sort: Robust associations multi-pedestrian tracking,'' \emph{arXiv preprint arXiv:2206.14651}, 2022.

\bibitem{deepocsort}
G.~Maggiolino, A.~Ahmad, J.~Cao, and K.~Kitani, ``Deep oc-sort: Multi-pedestrian tracking by adaptive re-identification,'' in \emph{2023 IEEE International conference on image processing (ICIP)}.\hskip 1em plus 0.5em minus 0.4em\relax IEEE, 2023, pp. 3025--3029.

\bibitem{tracktrack}
K.~Shim, K.~Ko, Y.~Yang, and C.~Kim, ``Focusing on tracks for online multi-object tracking,'' in \emph{Proceedings of the Computer Vision and Pattern Recognition Conference}, 2025, pp. 11\,687--11\,696.

\bibitem{topic}
X.~Cao, Y.~Zheng, Y.~Yao, H.~Qin, X.~Cao, and S.~Guo, ``Topic: A parallel association paradigm for multi-object tracking under complex motions and diverse scenes,'' \emph{IEEE Transactions on Image Processing}, 2025.

\bibitem{li2025multi}
Y.~Li, J.~Cao, M.~Naseer, Y.~Zhu, J.~Sun, Y.~Zhang, and F.~S. Khan, ``Multi-granularity language-guided training for multi-object tracking,'' \emph{IEEE Transactions on Circuits and Systems for Video Technology}, 2025.

\bibitem{ren2025semtg}
K.~Ren, C.~Hu, H.~Xi, Y.~Li, J.~Fan, and L.~Liu, ``Semtg-track: Multimodal fine-grained semantic-unit temporal guidance for multi-object tracking,'' \emph{Expert Systems with Applications}, p. 128359, 2025.

\bibitem{li2025dynamic}
X.~Li, B.~Zhong, Q.~Liang, Z.~Mo, J.~Nong, and S.~Song, ``Dynamic updates for language adaptation in visual-language tracking,'' in \emph{Proceedings of the Computer Vision and Pattern Recognition Conference}, 2025, pp. 19\,165--19\,174.

\bibitem{hurst2024gpt}
A.~Hurst, A.~Lerer, A.~P. Goucher, A.~Perelman, A.~Ramesh, A.~Clark, A.~Ostrow, A.~Welihinda, A.~Hayes, A.~Radford \emph{et~al.}, ``Gpt-4o system card,'' \emph{arXiv preprint arXiv:2410.21276}, 2024.

\bibitem{clip}
A.~Radford, J.~W. Kim, C.~Hallacy, A.~Ramesh, G.~Goh, S.~Agarwal, G.~Sastry, A.~Askell, P.~Mishkin, J.~Clark \emph{et~al.}, ``Learning transferable visual models from natural language supervision,'' in \emph{International conference on machine learning}.\hskip 1em plus 0.5em minus 0.4em\relax PmLR, 2021, pp. 8748--8763.

\bibitem{gal2022image}
R.~Gal, Y.~Alaluf, Y.~Atzmon, O.~Patashnik, A.~H. Bermano, G.~Chechik, and D.~Cohen-Or, ``An image is worth one word: Personalizing text-to-image generation using textual inversion,'' \emph{arXiv preprint arXiv:2208.01618}, 2022.

\bibitem{munkres1957algorithms}
J.~Munkres, ``Algorithms for the assignment and transportation problems,'' \emph{Journal of the society for industrial and applied mathematics}, vol.~5, no.~1, pp. 32--38, 1957.

\bibitem{bytetrack}
Y.~Zhang, P.~Sun, Y.~Jiang, D.~Yu, F.~Weng, Z.~Yuan, P.~Luo, W.~Liu, and X.~Wang, ``Bytetrack: Multi-object tracking by associating every detection box,'' in \emph{European conference on computer vision}.\hskip 1em plus 0.5em minus 0.4em\relax Springer, 2022, pp. 1--21.

\bibitem{hybrid}
M.~Yang, G.~Han, B.~Yan, W.~Zhang, J.~Qi, H.~Lu, and D.~Wang, ``Hybrid-sort: Weak cues matter for online multi-object tracking,'' in \emph{Proceedings of the AAAI conference on artificial intelligence}, vol.~38, no.~7, 2024, pp. 6504--6512.

\bibitem{coop}
K.~Zhou, J.~Yang, C.~C. Loy, and Z.~Liu, ``Learning to prompt for vision-language models,'' \emph{International Journal of Computer Vision}, vol. 130, no.~9, pp. 2337--2348, 2022.

\bibitem{cocoop}
------, ``Conditional prompt learning for vision-language models,'' in \emph{Proceedings of the IEEE/CVF conference on computer vision and pattern recognition}, 2022, pp. 16\,816--16\,825.

\bibitem{provp}
C.~Xu, Y.~Zhu, H.~Shen, B.~Chen, Y.~Liao, X.~Chen, and L.~Wang, ``Progressive visual prompt learning with contrastive feature re-formation,'' \emph{International Journal of Computer Vision}, vol. 133, no.~2, pp. 511--526, 2025.

\bibitem{zhang2024concept}
Y.~Zhang, C.~Zhang, K.~Yu, Y.~Tang, and Z.~He, ``Concept-guided prompt learning for generalization in vision-language models,'' in \emph{Proceedings of the AAAI Conference on Artificial Intelligence}, vol.~38, no.~7, 2024, pp. 7377--7386.

\bibitem{wei2022chain}
J.~Wei, X.~Wang, D.~Schuurmans, M.~Bosma, F.~Xia, E.~Chi, Q.~V. Le, D.~Zhou \emph{et~al.}, ``Chain-of-thought prompting elicits reasoning in large language models,'' \emph{Advances in neural information processing systems}, vol.~35, pp. 24\,824--24\,837, 2022.

\bibitem{wu2023referring}
D.~Wu, W.~Han, T.~Wang, X.~Dong, X.~Zhang, and J.~Shen, ``Referring multi-object tracking,'' in \emph{Proceedings of the IEEE/CVF conference on computer vision and pattern recognition}, 2023, pp. 14\,633--14\,642.

\bibitem{shao2024context}
Y.~Shao, S.~He, Q.~Ye, Y.~Feng, W.~Luo, and J.~Chen, ``Context-aware integration of language and visual references for natural language tracking,'' in \emph{Proceedings of the IEEE/CVF Conference on Computer Vision and Pattern Recognition}, 2024, pp. 19\,208--19\,217.

\bibitem{anh2024tp}
D.~L.~D. Anh, K.~H. Tran, and N.~H. Le, ``Tp-gmot: Tracking generic multiple object by textual prompt with motion-appearance cost (mac) sort,'' \emph{arXiv preprint arXiv:2409.02490}, 2024.

\bibitem{nguyen2023type}
P.~Nguyen, K.~G. Quach, K.~Kitani, and K.~Luu, ``Type-to-track: Retrieve any object via prompt-based tracking,'' \emph{Advances in Neural Information Processing Systems}, vol.~36, pp. 3205--3219, 2023.

\bibitem{feng2024memvlt}
X.~Feng, X.~Li, S.~Hu, D.~Zhang, J.~Zhang, X.~Chen, K.~Huang \emph{et~al.}, ``Memvlt: Vision-language tracking with adaptive memory-based prompts,'' \emph{Advances in Neural Information Processing Systems}, vol.~37, pp. 14\,903--14\,933, 2024.

\bibitem{ma2024unifying}
Y.~Ma, Y.~Tang, W.~Yang, T.~Zhang, J.~Zhang, and M.~Kang, ``Unifying visual and vision-language tracking via contrastive learning,'' in \emph{Proceedings of the AAAI Conference on Artificial Intelligence}, vol.~38, no.~5, 2024, pp. 4107--4116.

\bibitem{zgmot}
K.~H. Tran, A.~D.~L. Dinh, T.~P. Nguyen, T.~Phan, P.~Nguyen, K.~Luu, D.~Adjeroh, G.~Doretto, and N.~H. Le, ``Z-gmot: Zero-shot generic multiple object tracking,'' \emph{arXiv preprint arXiv:2305.17648}, 2023.

\bibitem{yu2023generalizing}
E.~Yu, S.~Liu, Z.~Li, J.~Yang, Z.~Li, S.~Han, and W.~Tao, ``Generalizing multiple object tracking to unseen domains by introducing natural language representation,'' in \emph{Proceedings of the AAAI Conference on Artificial Intelligence}, vol.~37, no.~3, 2023, pp. 3304--3312.

\bibitem{ipmot}
R.~Luo, Z.~Song, L.~Chen, Y.~Li, M.~Yang, and W.~Yang, ``Ip-mot: Instance prompt learning for cross-domain multi-object tracking,'' \emph{arXiv preprint arXiv:2410.23907}, 2024.

\bibitem{CAMOT}
F.~Limanta, K.~Uto, and K.~Shinoda, ``Camot: Camera angle-aware multi-object tracking,'' in \emph{Proceedings of the IEEE/CVF Winter Conference on Applications of Computer Vision}, 2024, pp. 6479--6488.

\bibitem{zhang2025robust}
Y.~Zhang, S.~Wang, Z.~Fu, L.~Zhao, and J.~Zhao, ``Robust multi-object tracking with pseudo-information guided motion and enhanced semantic vision,'' \emph{Expert Systems with Applications}, vol. 273, p. 126846, 2025.

\bibitem{jiang2023cross}
D.~Jiang and M.~Ye, ``Cross-modal implicit relation reasoning and aligning for text-to-image person retrieval,'' in \emph{Proceedings of the IEEE/CVF conference on computer vision and pattern recognition}, 2023, pp. 2787--2797.

\bibitem{mot17}
A.~Milan, L.~Leal-Taix{\'e}, I.~Reid, S.~Roth, and K.~Schindler, ``Mot16: A benchmark for multi-object tracking,'' \emph{arXiv preprint arXiv:1603.00831}, 2016.

\bibitem{mot20}
P.~Dendorfer, H.~Rezatofighi, A.~Milan, J.~Shi, D.~Cremers, I.~Reid, S.~Roth, K.~Schindler, and L.~Leal-Taix{\'e}, ``Mot20: A benchmark for multi object tracking in crowded scenes,'' \emph{arXiv preprint arXiv:2003.09003}, 2020.

\bibitem{sun2022dancetrack}
P.~Sun, J.~Cao, Y.~Jiang, Z.~Yuan, S.~Bai, K.~Kitani, and P.~Luo, ``Dancetrack: Multi-object tracking in uniform appearance and diverse motion,'' in \emph{Proceedings of the IEEE/CVF conference on computer vision and pattern recognition}, 2022, pp. 20\,993--21\,002.

\bibitem{hota}
J.~Luiten, A.~Osep, P.~Dendorfer, P.~Torr, A.~Geiger, L.~Leal-Taix{\'e}, and B.~Leibe, ``Hota: A higher order metric for evaluating multi-object tracking,'' \emph{International journal of computer vision}, vol. 129, no.~2, pp. 548--578, 2021.

\bibitem{idf1}
E.~Ristani, F.~Solera, R.~Zou, R.~Cucchiara, and C.~Tomasi, ``Performance measures and a data set for multi-target, multi-camera tracking,'' in \emph{European conference on computer vision}.\hskip 1em plus 0.5em minus 0.4em\relax Springer, 2016, pp. 17--35.

\bibitem{mota}
K.~Bernardin and R.~Stiefelhagen, ``Evaluating multiple object tracking performance: the clear mot metrics,'' \emph{EURASIP Journal on Image and Video Processing}, vol. 2008, no.~1, p. 246309, 2008.

\bibitem{li2023clip}
S.~Li, L.~Sun, and Q.~Li, ``Clip-reid: exploiting vision-language model for image re-identification without concrete text labels,'' in \emph{Proceedings of the AAAI conference on artificial intelligence}, vol.~37, no.~1, 2023, pp. 1405--1413.

\bibitem{zeng2022motr}
F.~Zeng, B.~Dong, Y.~Zhang, T.~Wang, X.~Zhang, and Y.~Wei, ``Motr: End-to-end multiple-object tracking with transformer,'' in \emph{European conference on computer vision}.\hskip 1em plus 0.5em minus 0.4em\relax Springer, 2022, pp. 659--675.

\bibitem{cao2025topic}
X.~Cao, Y.~Zheng, Y.~Yao, H.~Qin, X.~Cao, and S.~Guo, ``Topic: a parallel association paradigm for multi-object tracking under complex motions and diverse scenes,'' \emph{IEEE Transactions on Image Processing}, 2025.

\end{thebibliography}
\begin{IEEEbiography}[{\includegraphics[width=1in,height=1.25in,clip,keepaspectratio]{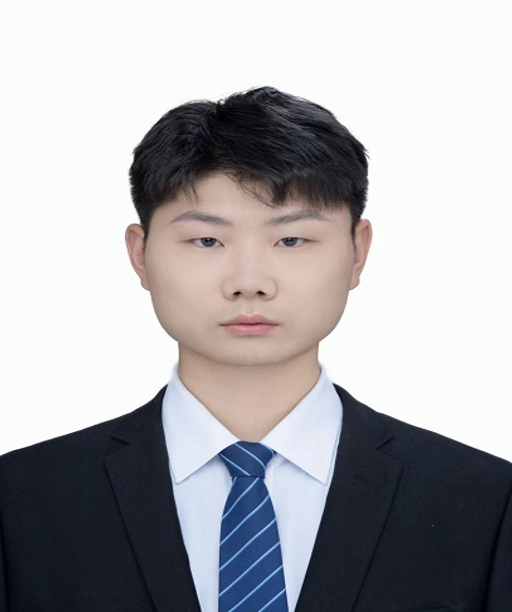}}]{Yukuan Zhang}
was born in 1997. He received the M.S. degree from the School of Physics and Electronic Information, Yunnan Normal University in 2024. He is currently a Ph.D. candidate with the College of Computer Science and Technology, Jilin University, and his supervisor is Professor Shengsheng Wang. His research interests include computer vision, multi-object tracking, multi-modal fusion, and prompt learning.
\end{IEEEbiography}
\vspace{0pt}
\begin{IEEEbiography}[{\includegraphics[width=1in,height=1.25in,clip,keepaspectratio]{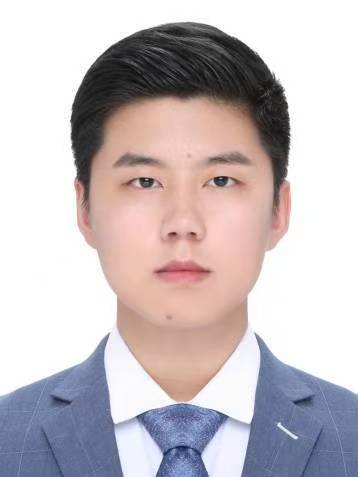}}]{Jiarui Zhao}
was born in 2002. He earned his B.S. degree in 2024 from the College of Software, Jilin University, China, where he is currently a Master’s candidate in Software Engineering, and his supervisor is Professor Shengsheng Wang. His research interests encompass multi-target object tracking, computer vision, and related machine learning techniques.
\end{IEEEbiography}
\vspace{0pt}
\begin{IEEEbiography}[{\includegraphics[width=1in,height=1.25in,clip,keepaspectratio]{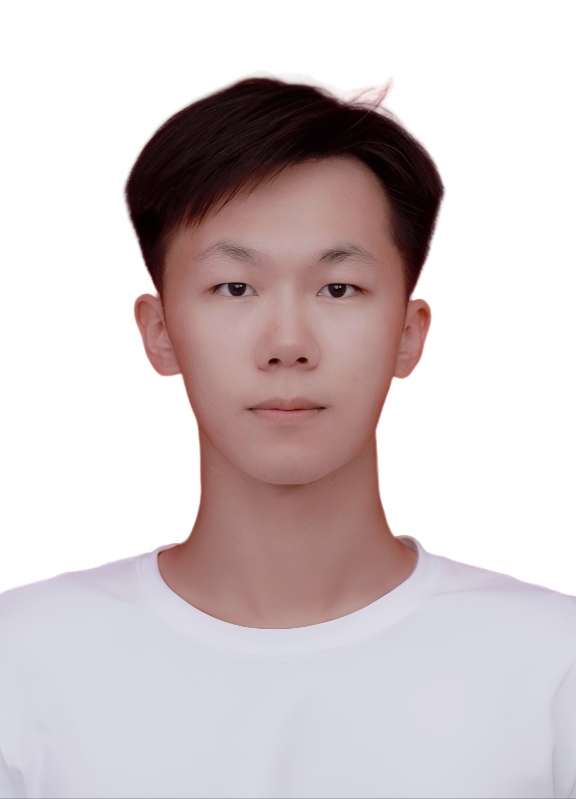}}]{Shangqing Nie}
received his Bachelor's degree from Shandong University of Science and Technology in 2025. He is currently a Master's student at the College of Software, Jilin University. His main research interest is visible-infrared multi-object tracking.
\end{IEEEbiography}
\vspace{0pt}
\begin{IEEEbiography}[{\includegraphics[width=1in,height=1.25in,clip,keepaspectratio]{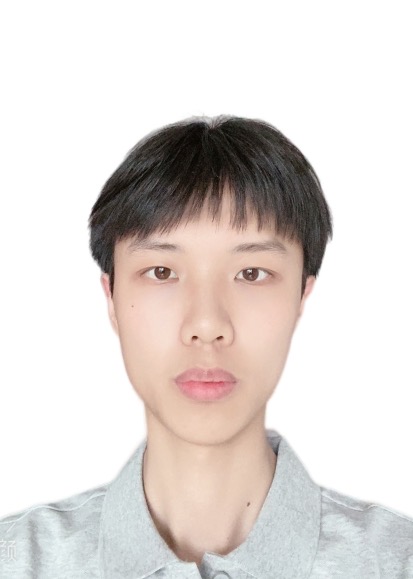}}]{Jin Kuang}
was born in 2001. He received the B.S. degree from Xiangnan University in 2022. He is currently pursuing the M.S. degree with Yangtze University, Wuhan, China, and serves as a Research Assistant with the Hunan Engineering Research Center of Advanced Embedded Computing and Intelligent Medical Systems. His research interests include domain adaptation, image segmentation, and low-light image enhancement.\end{IEEEbiography}
\vspace{0pt}
\begin{IEEEbiography}[{\includegraphics[width=1in,height=1.25in,clip,keepaspectratio]{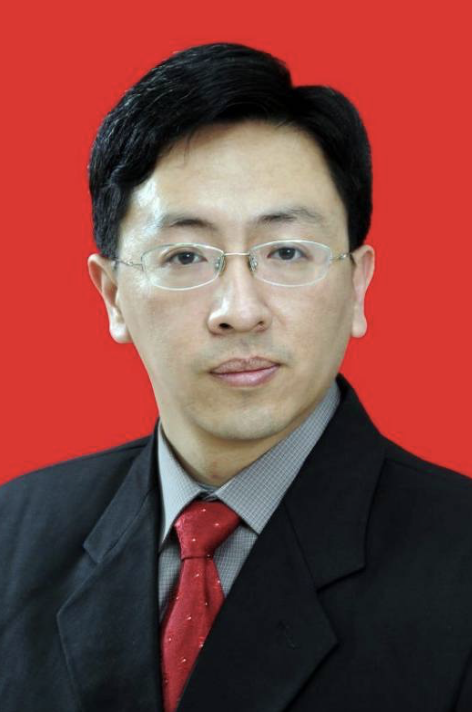}}]{Shengsheng Wang}
received the B.S., M.S., and Ph.D. degrees in computer science from Jilin University, Changchun, China, in 1997, 2000, and 2003, respectively. He is currently a Professor with the College of Computer Science and Technology, Jilin University. His research interests are in the areas of computer vision, deep learning, and data mining.
\end{IEEEbiography}
\vfill

\end{document}